\documentclass[10pt,twocolumn,letterpaper]{article}

\usepackage{cvpr}              % To produce the CAMERA-READY version
\usepackage{graphicx}
\usepackage{amsmath}
\usepackage{amssymb}
\usepackage{booktabs}
\usepackage{multirow}
\usepackage[table]{xcolor}
\usepackage{subcaption}

 \usepackage[pagebackref,breaklinks,colorlinks,allcolors=cvprblue]{hyperref}
\newtheorem{theorem}{Theorem}[section] % 按章节编号
\usepackage[capitalize]{cleveref}
\crefname{section}{Sec.}{Secs.}
\Crefname{section}{Section}{Sections}
\Crefname{table}{Table}{Tables}
\crefname{table}{Tab.}{Tabs.}

\definecolor{lightblue}{RGB}{226,240,255} % 柔和浅蓝

\usepackage{booktabs}            % beautiful rules
\usepackage[table]{xcolor}       % \rowcolor etc.
\usepackage{siunitx}             % decimal alignment
\usepackage{multirow,makecell}   % stacked cells / headers
\usepackage{array}               % custom column types
\usepackage{adjustbox}           % fit to width if needed

\renewcommand{\arraystretch}{1.05}

\newcommand{\Best}[1]{\textbf{#1}}
\newcommand{\dlt}[1]{\textcolor{blue}{\scriptsize #1}} % delta in blue

\definecolor{cvprblue}{rgb}{0.21,0.49,0.74}

\def\paperID{10548} % *** Enter the Paper ID here
\def\confName{CVPR}
\def\confYear{2026}

\title{Fixed Anchors Are Not Enough: Dynamic Retrieval and Persistent Homology for Dataset Distillation}

\author{Muquan~Li, Hang~Gou, Yingyi~Ma, Rongzheng~Wang, Ke~Qin, Tao~He\thanks{Corresponding author.}\\
The Laboratory of Intelligent Collaborative Computing of UESTC\\
{\tt\small \{muquanli2023,gouhang,mayingyi,wangrongzheng\}@std.uestc.edu.cn}\\
{\tt\small qinke@uestc.edu.cn, tao.he01@hotmail.com}
}

\begin{document}
\maketitle
\begin{abstract}
Decoupled dataset distillation (DD) compresses large corpora into a few synthetic images by matching a frozen teacher’s statistics. However, current residual-matching pipelines rely on static real patches, creating a fit-complexity gap and a pull-to-anchor effect that reduce intra-class diversity and hurt generalization. To address these issues,  we introduce \textbf{RETA}---a \textbf{Re}trieval and \textbf{T}opology \textbf{A}lignment framework for decoupled DD. First, \textbf{Dynamic Retrieval Connection (DRC)} selects a real patch from a prebuilt pool by minimizing a fit-complexity score in teacher feature space; the chosen patch is injected via a residual connection to tighten feature fit while controlling injected complexity. Second, \textbf{Persistent Topology Alignment (PTA)} regularizes synthesis with persistent homology: we build a mutual k-NN feature graph, compute persistence images of components and loops, and penalize topology discrepancies between real and synthetic sets, mitigating pull-to-anchor effect. Across CIFAR-100, Tiny-ImageNet, ImageNet-1K, and multiple ImageNet subsets, RETA consistently outperforms various baselines under comparable time and memory, especially reaching 64.3\% top-1 accuracy on ImageNet-1K with ResNet-18 at 50 images per class, +3.1\% over the best prior. 
\end{abstract}

% \begin{abstract}
% Decoupled dataset distillation offers a scalable way to compress large-scale datasets into a few synthetic images by matching global statistics of a frozen teacher model. However, existing pipelines rely on static real patches, which induce a fit–complexity gap and a pull-to-anchor effect that limit generalization. To address these limitations, we propose Retrieval and Topology Alignment (RETA), a decoupled DD framework based on residual matching with topology regularization. Specifically, we first introduces a Dynamic Retrieval Connection (DRC) that retrieves a real patch from a pre-built pool by minimizing a fit–complexity score in a frozen teacher model space, explicitly shrinking the fit gap while penalizing high complexity. Additionally, we introduce a Persistent Topology Alignment (PTA), which applies persistent homology on a mutual $k$-NN graph to align the multi-scale connectivity and loop statistics of synthetic and real features, mitigating the pull-to-anchor effect. Extensive experiments on CIFAR-100, Tiny-ImageNet, ImageNet-1K and multiple ImageNet subsets demonstrate that RETA consistently surpasses current decoupled baselines, especially achieving up to 3.1\% improvement on ImageNet-1K using ResNet18 with 50 images per class, while remaining competitive in time and memory. 
% \end{abstract}    
\section{Introduction}
\label{sec:intro}

\begin{figure}[t]
\centering
\includegraphics[width=0.97\columnwidth]{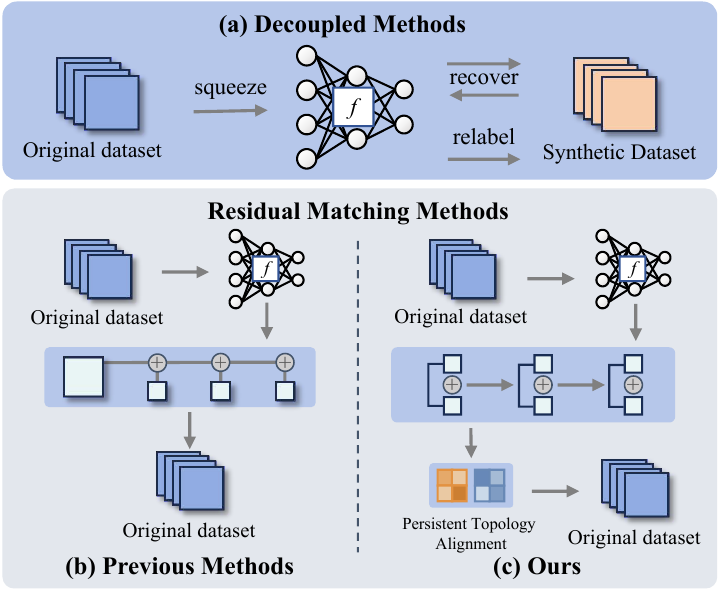}
\caption{Comparison between existing pipelines and our proposed RETA. \textbf{(a)} Decoupled methods squeeze the original dataset into a compact set. \textbf{(b)} Previous residual matching methods \cite{FADRM} connect a fixed real patch to the distilled images at every step. \textbf{(c)} RETA augments residual matching with Dynamic Retrieval Connection (DRC) and Persistent Topology Alignment (PTA).}
\label{fig:intro}
\end{figure}

% The rapid scaling of large language and vision models has created an unprecedented demand for data, rendering repeated training and data governance increasingly costly \cite{DeepLearning_1, Survey_2}. Dataset Distillation \cite{DatasetDistillation,Survey_1} (DD) tackles this by compressing a large dataset into a small synthetic dataset that yields comparable downstream performance when used for training. It is typically realized in bilevel optimization, so that a few steps on synthetic dataset mimic many steps on original dataset, and the same synthetic dataset can be reused across architectures and training schedules. However, optimizing supervision and distributional alignment within the bilevel updates entangles their gradients, leading to interference and poor performance \cite{Survey_1, SRe2L}.

Recent advances in deep learning models \cite{naveed2025comprehensive,wang2026rethinking,He2022OpenVocabSGG,ke2025early,UnifiedTransformerSGGHOI,tian2026stpe} have intensified the demand for high-quality data, rendering repeated full-dataset training and data governance increasingly costly \cite{DeepLearning_1, Survey_2,DBLP:conf/mm/LiZ0XLQ24,erhc,Xie_2024_CVPR,cdd,xie2024distillation,xie2024pairwise,he2025prism,he2026monotonic,hu2025spade,he2026lifelong}. Dataset Distillation (DD) \cite{DatasetDistillation,Survey_1,DBLP:conf/aaai/LiZDXQ25,libeyond} seeks to mitigate this by synthesizing a compact surrogate dataset that, when used for training, matches the performance of the original dataset. Most DD \cite{DataMatching_MTT, DataMatching_GM} formulate synthesis as bilevel optimization: the synthetic set is updated so that a few gradient steps on it mimic many steps on real data. However, jointly enforcing supervision and distributional alignment within the bilevel updates entangles their gradients, causing interference, unstable dynamics, and degraded performance \cite{Survey_1, SRe2L}. This motivates approaches that explicitly decouple supervision from alignment to stabilize optimization and improve generalization.

Decoupled DD \cite{SRe2L, G-VBSM,EDC, CV-DD, DELT, NRR-DD,FADRM}  separates the supervised objective from distributional alignment into two optimization streams, enabling independent control over task loss and data matching \cite{DataMatching_GM,DataMatching_MTT,libeyond,li2026efficient,ma2026efficient}. This design yields practical benefits: the supervised learner remains architecture-agnostic and plug-and-play across backbones and schedules, while the alignment module can leverage pretrained representations and stronger metrics, improving stability and transfer. Building on this perspective, residual-matching \cite{FADRM} approaches periodically anchor a synthetic image to the real-data manifold by additively injecting a small real patch through a residual connection, rather than relying solely on gradient-based updates, as shown in Fig.~\ref{fig:intro}. The anchor mitigates pixel-space drift \cite{DBLP:conf/icml/YangZDR24} during synthesis and preserves fine-scale, high-frequency structure.

% Recently, decoupled DD \cite{SRe2L, EDC, CV-DD} optimizes the supervised objective and the distributional alignment in separate streams, allowing each to be controlled independently. This separation confers several advantages: the supervised term remains architecture-agnostic and plug-and-play, while the alignment term can exploit pretrained features, improving stability and transfer. Building on this, residual matching \cite{FADRM} schemes have subsequently emerged. Instead of relying solely on gradient updates, they periodically anchor a synthetic image to the real data manifold by residual connecting a real patch. This simple mechanism curbs drift from pure pixel optimization and preserves high-frequency structures.

Despite strong gains \cite{FADRM}, we argue that prevailing residual-matching pipelines exhibit two coupled failure modes. \textbf{(i) Fit–complexity gap.} Using a fixed sampled residual patch can misalign with the current synthetic features, enlarging the fit gap; conversely, injecting a highly textured patch can inflate hypothesis complexity, loosening the post-connection generalization bound. \textbf{(ii) Pull-to-anchor effect.} Repeatedly anchoring to teacher-nearest real samples contracts intra-class synthetic features, causing early cluster merging in the learned representation. Accumulating across stages, these effects erode intra-class diversity and yield brittle decision boundaries. This raises a research question: \emph{Can we design an adaptive residual-matching mechanism that simultaneously balances feature fit and complexity and preserves class diversity?}

% Despite their promise \cite{FADRM}, current residual matching pipelines incur two limitations. First, a fixed sampled residual patch introduces a fit–complexity gap: a patch that poorly matches the current synthetic features widens the feature fit gap, whereas a highly textured patch may raise the hypothesis complexity, loosening generalization bounds after the connection. Second, repeatedly anchoring to nearby real samples in the teacher feature space produces a pull-to-anchor effect, where intra-class synthetic features contract too aggressively, causing early cluster merging in the learned representation. These effects accumulate across stages, reducing diversity within classes and leaving decision boundaries brittle.

We thus propose \textbf{RETA}—a \underline{Re}trieval and \underline{T}opology \underline{A}lignment framework that operates within decoupled DD. RETA comprises two components. \textbf{(i) Dynamic Retrieval Connection (DRC)}: to reduce the fit gap while controlling hypothesis complexity, we prebuild per-class pools of image-level ($1\times1$-grid) real patches. At each residual step, a frozen teacher embeds the current synthetic and all candidates; we select the patch that minimizes a fit–complexity score that trades fit gap against a complexity gap. The selected patch is resampled to the current resolution and injected via a residual connection, yielding anchors that are both feature-aligned and low-complexity. \textbf{(ii) Persistent Topology Alignment (PTA)}: a differentiable persistent-homology regularizer \cite{PH} that constructs a \(k\)-NN graph in teacher space, computes persistence images capturing how connected components and loops evolve with radius, and penalizes discrepancies between real and synthetic topological signatures. Together, DRC tightens fit-complexity gap while PTA preserves class topology, mitigating pull-to-anchor effect and producing synthetic data that are geometrically faithful to the real distribution. Under comparable compute, RETA consistently improves accuracy and generalization over various DD baselines.

In summary, our contributions are threefold as below:
\begin{itemize}
\item \textbf{Analysis.} We theoretically characterize two failure modes in decoupled DD with residual matching, the {fit–complexity gap} and a {pull-to-anchor effect}, that erode intra-class diversity and weaken generalization.
\item \textbf{Method.} We introduce \textbf{RETA}, comprising (i) {Dynamic Retrieval Connection} (DRC), which retrieves real patches via a fit–complexity score and injects them  to tighten feature fit while controlling injected complexity; and (ii) {Persistent Topology Alignment} (PTA), a persistent-homology regularizer that matches persistence images to preserve class topology and counter pull-to-anchor effects.
\item \textbf{Results.} RETA achieves state-of-the-art (SOTA) performance across multiple architectures and datasets; e.g., on ImageNet-1K (IPC=$50$) with ResNet-$18$, RETA attains $64.3$\% accuracy, surpassing the best prior \cite{FADRM} by $3.1$\%.
\end{itemize}

% \begin{itemize}
%     \item We theoretically identify two failure modes in decoupled DD with residual matching: the fit–complexity gap and a pull-to-anchor effect.
%     \item We introduce Dynamic Retrieval Connection (DRC), a retrieval in the same class that selects real patches via a fit–complexity score to shrink the fit gap while limiting injected complexity.
%     \item We propose Persistent Topology Alignment (PTA), a differentiable alignment by persistent homology that matches persistence images to preserve intra-class topology and counter pull-to-anchor effect.
%     \item Extensive experiments show that RETA achieves state-of-the-art (SOTA) performance across all evaluated architectures and datasets. For instance, on ImageNet-1K \cite{ImageNet-1K} with IPC=50 using ResNet18 \cite{ResNet18}, RETA attains an accuracy of 64.3\%, surpassing the SOTA baseline by 3.1\%. 
% \end{itemize}
\section{Related work}

\noindent\textbf{Dataset Distillation.}
Dataset Distillation (DD)~\cite{DatasetDistillation,Survey_1} aims to synthesize a small set of examples that preserves the training behavior of a much larger corpus. Most prior work casts DD as a bilevel optimization~\cite{DatasetDistillation,qiu2025convex}, where synthetic data are updated so that a model trained on them mimics a model trained on real data, i.e., an approach that incurs expensive nested optimization and large memory footprints~\cite{Survey_2}. Within this paradigm, three main families have emerged: \emph{gradient matching}~\cite{DataMatching_GM,DataMatching_DSA}, which aligns per-batch gradients computed on real versus synthetic data; \emph{trajectory matching}~\cite{DataMatching_MTT,DataMatching_FTD,DataMatching_DATM}, which enforces agreement along training trajectories; and \emph{distribution matching}~\cite{DataMatching_DM,DataMatching_NCFM}, which aligns feature statistics across network layers, often using a pretrained teacher. While effective on moderate-scale benchmarks, the bilevel nature of these objectives leads to substantial computational overhead and limits scalability to large, high-resolution datasets~\cite{Survey_2,Survey_3}.

% \noindent\textbf{Dataset Distillation.}
% Dataset Distillation (DD) \cite{DatasetDistillation,Survey_1} aims to synthesize a compact dataset that captures the essential information of a large original dataset. Early DD methods are typically formulated as bilevel optimization problems \cite{DatasetDistillation}, which seek to align data properties between real and synthetic datasets through computationally intensive nested optimization loops \cite{Survey_2}. Representative approaches include gradient matching \cite{DataMatching_GM, DataMatching_DSA}, which minimizes the discrepancy between gradients computed on real and synthetic data batches; trajectory matching \cite{DataMatching_MTT, DataMatching_FTD, DataMatching_DATM}, which aligns the full optimization trajectories of models trained on the two datasets; and distribution matching \cite{DataMatching_DM, DataMatching_NCFM}, which aligns feature distributions of real and synthetic data across multiple layers of a pretrained network. Despite their effectiveness, these methods suffer from high computational costs and limited scalability to large-scale datasets due to their bilevel optimization structure \cite{Survey_2,Survey_3}.

\noindent\textbf{Decoupled Optimization.}
To mitigate the inefficiency of bilevel DD, recent work adopts decoupled objectives that optimize synthetic data by matching global statistics of a pretrained teacher, thereby removing costly inner-loop retraining~\cite{SRe2L,G-VBSM,RDED}. SRe$^2$L~\cite{SRe2L} first demonstrated the effectiveness of aligning teacher-driven model statistics; subsequent variants broaden this design space: EDC stabilizes optimization by initializing from real image patches~\cite{EDC}, and CV-DD steers synthesis by aligning BatchNorm population statistics~\cite{CV-DD}. More recently, FADRM~\cite{FADRM} identified an {information vanishing} failure mode and proposed an Adjustable Residual Connection that periodically injects features from preselected real images to preserve signal. However, this static connection induces two biases, a {fit–complexity gap} and a {pull-to-anchor} effect, that tether synthesis to fixed, high-complexity anchors and ultimately limit the quality and generalization of the distilled data.

% \textbf{Uni-Level Optimization.}
% To address the efficiency limitations of bilevel optimization, a recent line of work has introduced uni-level optimization frameworks. These methods bypass the costly nested-loop optimization by matching global statistics from pre-trained teacher models \cite{SRe2L,G-VBSM, RDED}. Seminal approaches such as SRe²L \cite{SRe2L} demonstrated the effectiveness of aligning such model statistics. This direction has been rapidly advanced by methods like EDC \cite{EDC}, which initializes synthetic data from real image patches, and CV-DD \cite{CV-DD}, which focuses on aligning global BatchNorm statistics. More recently, FADRM \cite{FADRM} further explored this framework and identified a critical information vanishing problem. To mitigate this issue, FADRM proposed an Adjustable Residual Connection \cite{FADRM} that periodically injects features from preselected real images. However, this static connection mechanism introduces a fit-complexity gap and a pull-to-anchor effect, which ultimately constrain the quality and generalization capability of the distilled dataset.

\section{Preliminary}
\label{subsec:prelim}

Given a large dataset $\mathcal{D}=\{(x_i,y_i)\}_{i=1}^{|\mathcal{D}|}$ and a student $f_\theta$, decoupled dataset distillation \cite{SRe2L, EDC, CV-DD, FADRM} aims to learn a compact synthetic dataset $\tilde{\mathcal{C}}=\{(\tilde{x}_j,\tilde{y}_j)\}_{j=1}^N$ ($N\ll|\mathcal{D}|$) by decoupling two objectives: a supervised objective on the synthetic set and a distributional alignment to the real data. Its formulation with teacher $\mathcal{T}$ is defined as:
\begin{equation}
\label{eqn:decoupled}
\min_{\tilde{\mathcal{C}}}\;\; 
\mathcal{L}_{\text{sup}}\big(f_\theta;\tilde{\mathcal{C}}\big)
\;+\;
\beta\,\mathcal{R}_{\text{align}}\big(\tilde{\mathcal{C}};\mathcal{D},\mathcal{T}\big),
\end{equation}
where $\mathcal{L}_{\text{sup}}$ denotes typically a cross-entropy loss computed on $\tilde{\mathcal{C}}$, $\mathcal{R}_{\text{align}}$ aligns global statistics of $\tilde{\mathcal{C}}$ to those of $\mathcal{D}$, and $\beta>0$ balances the two terms. However, decoupled DD  directly optimizes pixels of $\tilde{x}$ by gradient descent on Eqn.~\ref{eqn:decoupled}, without simulating long training trajectories \cite{DataMatching_MTT} on $\mathcal{D}$. This is prone to information vanishing \cite{FADRM}: purely pixel updates under repeated resampling gradually wash out fine structures that are hard to recover by the alignment term alone.

Recent work (e.g., FADRM \cite{FADRM}) addresses this problem with a residual matching that periodically anchors the synthetic image to the real-data manifold. Specifically, it partitions the total optimization budget $B$ into $k{+}1$ blocks, each with $b=\lfloor B/(k{+}1)\rfloor$ steps on Eqn.~\ref{eqn:decoupled}. At the end of each of the first $k$ blocks, it applies a static residual connection:
\begin{equation}
\label{eqn:arc}
\tilde{x}_t \;\leftarrow\; \alpha\,\tilde{x}_t \;+\; (1-\alpha)\,\mathrm{Resample}(P, D_t),
\end{equation}
where $P$ denotes a real patch randomly selected from the original dataset, $\mathrm{Resample}$ denotes broadcast tiling to resolution $D_t$ at step $t$, and $\alpha\in[0,1]$ controls the connection ratio. Intuitively, residual matching in Eqn.~\ref{eqn:arc} forms a convex combination between the optimized synthetic image and a real sample, continuously injecting real structures that are otherwise eroded by pure pixel updates. Nevertheless, a fixed residual connection simultaneously induces a fit–complexity mismatch and a pull-to-anchor effect that collapses class geometry.

% ,height=0.36\textheight,keepaspectratio

\begin{figure*}[t]
\centering
\includegraphics[width=1.9\columnwidth]{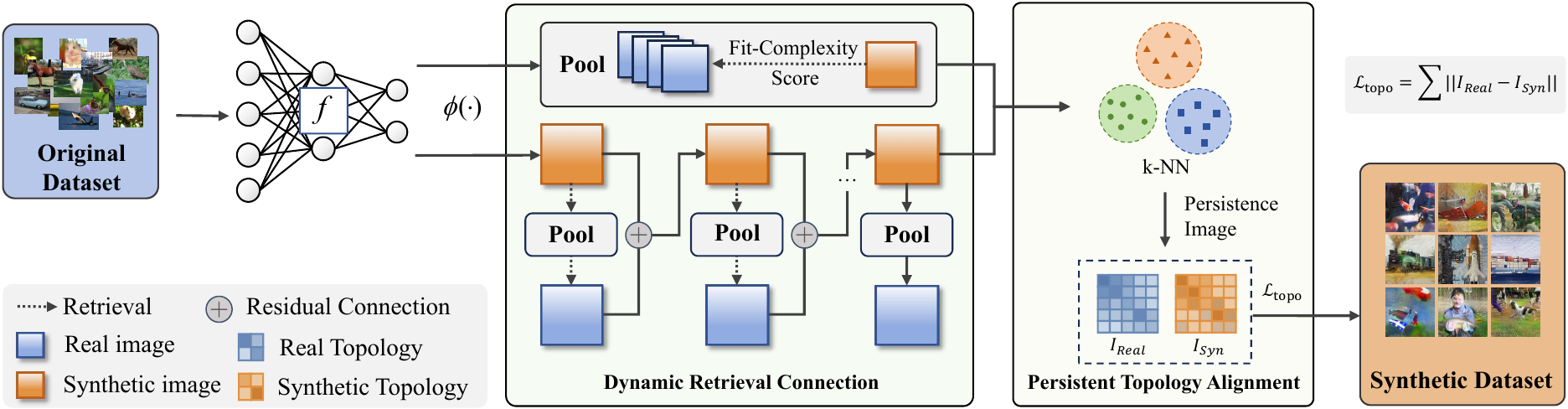} 
\caption{Overview of RETA. Dynamic Retrieval Connection (top) computes a fit–complexity score to adaptively retrieve real patches as anchors for residual matching, while Persistent Topology Alignment (bottom) aligns the persistent diagrams of real and synthetic features. The two modules are jointly optimized to produce synthetic datasets that retain both accuracy and topology-aware structure.} 
\label{fig:method}
\end{figure*}

\section{Method}  

% We propose \textbf{RETA} (Retrieval and Topology Alignment) for dataset distillation (Fig.~\ref{fig:method}). We first revisit the decoupled DD setup that optimizes synthetic images by \emph{residual matching} to a frozen teacher. Building on this base, we introduce a \emph{Dynamic Retrieval Connection} (DRC) in Sec. \S \tao{XX} that, at each optimization stage, retrieves a real patch from a curated pool by minimizing a teacher-space {fit–complexity} score. We then add \emph{Persistent Topology Alignment} (PTA) in Sec. \S \tao{XX}, a differentiable homology regularizer that matches the multi-scale connectivity and loop statistics of synthetic and real features via persistence images, preserving class geometry and mitigating the pull-to-anchor effect. Together, DRC and PTA stabilize optimization, prevent geometric collapse, and improve the fidelity and generalization of the distilled set.

% In this section, we detail our proposed Retrieval and Topology Alignment for Dataset Distillation (RETA) framework, as shown in Figure \ref{fig:method}. We first revisit decoupled dataset distillation and the residual matching used in prior work. Building on this setup, we introduce a Dynamic Retrieval Connection (DRC) that selects a real patch at each stage to simultaneously shrink the feature fit gap and control hypothesis complexity. We then present Persistent Topology Alignment (PTA), a homology regularizer that preserves class geometry and mitigats the pull-to-anchor effect.
\noindent\textbf{Motivation.} Our following analysis uncovers two coupled failure modes in residual-matching distillation. \emph{Locally}, each residual update in Eqn.~\ref{eqn:arc} must balance the fit of an injected real patch to the current synthetic image against the extra hypothesis complexity (\S \ref{subsec:dynarc}), i.e.,  naive anchors often narrow the fit gap while unnecessarily inflating complexity. \emph{Globally}, repeatedly mixing anchors with nearby teacher-space features contracts synthetic representations toward a few anchors, yielding a pull-to-anchor effect (\S \ref{subsec:topo}) in which intra-class clusters fuse prematurely and class topology collapses. These observations call for (i) a principled, stage-wise anchor selection that manages the fit–complexity trade-off, and (ii) a topology-aware constraint that preserves the evolving geometry in teacher space.

\noindent\textbf{Overview.} The overall architecture of \textbf{RETA} is illustrated in Fig.~\ref{fig:method}. RETA comprises two primary modules: \emph{Dynamic Retrieval Connection} (\textbf{DRC}) and \emph{Persistent Topology Alignment} (\textbf{PTA}). \textbf{DRC} replaces fixed anchors with per-stage, per-class retrieval from a curated pool of real patches, selecting anchors by minimizing a fit–complexity objective to reduce the fit gap while controlling hypothesis complexity (\S \ref{subsec:dynarc}). \textbf{PTA} regularizes global geometry in the teacher space by aligning persistence summaries of synthetic and real features, thereby preserving multi-scale connectivity and loop structure (\S \ref{subsec:topo}). Together, DRC determines \emph{which} real patches to connect at each stage, while PTA enforces \emph{what} global structure those connections should maintain.

\subsection{Dynamic Retrieval Connection}
\label{subsec:dynarc}

\noindent \textbf{Fit–Complexity Analysis.}
Residual connections \cite{FADRM} alter two coupled forces: the \emph{fit} of the connected real patch to the current synthetic image and the \emph{complexity} they inject into the hypothesis class. Following FADRM’s generalization analysis \cite{FADRM}, the post-connection risk can be decomposed into a data-dependent fit term and a capacity penalty that grows with the added complexity. Static, preselected patches may improve fit but often inflate complexity, widening a \emph{fit–complexity gap} that degrades generalization and encourages anchoring. This  motivates seeking connections that jointly optimize fit while controlling complexity, rather than relying on fixed anchors.

 \begin{theorem}\label{thm:sba}
Let $\mathfrak{R}_n(\cdot)$ denote the empirical Rademacher complexity and let $H\circ\mathcal{S}$ be the composition of a hypothesis class $H$ with a sample set $\mathcal{S}$. Consider a pre-connection synthetic set $\tilde{\mathcal{C}}_{\mathrm{pre}}$ and a post-connection set $\tilde{\mathcal{C}}_{\mathrm{post}}$ obtained by applying a residual connection with mixing weight $\alpha\in[0,1]$ to a set of real samples $O$. Then
\begin{equation}
\begin{aligned}
\mathfrak{R}_n\!\big(H\!\circ\!\tilde{\mathcal{C}}_{\text{post}}\big)
- \mathfrak{R}_n\!\big(H\!\circ\!\tilde{\mathcal{C}}_{\text{pre}}\big)
\;&\le\;  \\
(1-\alpha)\!\left[
\mathfrak{R}_n\!\big(H\!\circ\!O\big)
\!- \!\mathfrak{R}_n\!\big(H\!\circ\!\tilde{\mathcal{C}}_{\text{pre}}\big)
\right]& \!+\! L \alpha(1-\alpha)\,\Delta .
\label{eq:rad-bound}
\end{aligned}
\end{equation}
% \begin{equation}
% \label{eq:rad-bound}
% \mathfrak{R}_n\big(H\circ\tilde{\mathcal{C}}_{\mathrm{post}}\big)
% -\mathfrak{R}_n\big(H\circ\tilde{\mathcal{C}}_{\mathrm{pre}}\big)
% \;\le\;
% (1-\alpha)\left[\mathfrak{R}_n\big(H\circ\big)
% -\mathfrak{R}_n\big(H\circ\tilde{\mathcal{C}}_{\mathrm{pre}}\big)\right]
% \;+\; L\,\alpha(1-\alpha)\,\Delta,
% \end{equation}
where $L\!>\!0$ is a constant depending on $H$ and the connection operator, and
\(
\Delta \!=\!\frac{1}{|M|}\sum_{(i,P)\in M}\bigl\|\tilde{x}^{\mathrm{pre}}_{i}\!-\!P\bigr\|_2^{2}
\)
is the average \emph{fit gap} between each connected synthetic sample $\tilde{x}^{\mathrm{pre}}_{i}$ and its paired real patch $P\in O$ in a matching set $M$.
\end{theorem}
Please refer to FADRM~\cite{FADRM} for the proof. Eqn.~\ref{eq:rad-bound} highlights two levers for tightening the post-connection generalization bound: (i) \emph{shrink the fit gap} $\Delta$ by selecting real patches close to the current synthetic samples; and (ii) \emph{control the complexity gap} $\mathfrak{R}_n(H\circ)-\mathfrak{R}_n(H\circ\tilde{\mathcal{C}}_{\mathrm{pre}})$ by ensuring the participating real set $O$ is sufficiently regular. The first term reduces the bias introduced by residual connection, while the second prevents a capacity rebound from noisy anchors that inflate $\mathfrak{R}_n(H\circ)$. Consequently, a fixed anchor is suboptimal: it neither adapts across stages to minimize $\Delta$ nor reliably regularizes complexity. This motivates our stage-wise \emph{dynamic retrieval} of $P$ before each residual connection to jointly control both terms.

% Please refer to Appendix~\textcolor{red}{A} for the proof.

\paragraph{Formulation.}
Guided by the analysis above, we introduce a \emph{Dynamic Retrieval Connection} (DRC) that adapts the anchor $P$ at every stage. For each class $c$, we construct a pool $p_c$ containing one $1{\times}1$ patch (the entire image) per real image. A frozen encoder $\phi(\cdot)$ produces embeddings for both real patches and the current synthetic query. For any candidate patch $o\in p_c$, we define a \emph{fit-complexity score}:
\begin{equation}
J(o \mid \tilde{x}_t)
\;=\;
(1-\lambda)\,\big\| q(\tilde{x}_t) - z(o) \big\|_2^{2}
\;+\;
\lambda\, c(o),
\end{equation}
where \(q(\tilde{x}_t)=\mathrm{Norm}\!\big(\phi(\tilde{x}_t)\big),\;
z(o)=\mathrm{Norm}\!\big(\phi(o)\big)\) and  $\lambda\in[0,1]$ trades off fit and complexity. The first term pulls the retrieved patch toward the current synthetic feature, directly contracting the fit gap $\Delta$. To control complexity, we penalize patches with irregular high-frequency content via
\begin{equation}
    c(o)
\;=\;
\mathrm{Var}_{u\in\Omega_{D_t}}\!\Big(\,\big\|\nabla\!\big(G_\sigma * o\big)(u)\big\|_2^{2}\,\Big) ,
\end{equation}
with $*$ denoting spatial convolution and $\Omega_{D_t}$ the $D_t{\times}D_t$ pixel grid. Intuitively, large $c(o)$ indicates strong spatial fluctuation in gradient magnitude even after Gaussian smoothing (i.e., residual sharp edges); fitting such content increases the predictor’s effective smoothness which inflates $\mathfrak{R}_n(H\!\circ\!\cdot)$ by norm-based generalization bounds.

We retrieve \(o^\star = \arg\min_{o\in p_c} J(o \mid \tilde{x}_t),
\) resample it to the current resolution, and apply the residual update:
\begin{equation}\label{eqn:drc}
\tilde{x}_t \;\leftarrow\; \alpha\,\tilde{x}_t \;+\; (1-\alpha)\,\mathrm{Resample}(o^\star, D_t).
\end{equation}
Unlike a fixed anchor, DRC operationalizes the two levers in Theorem~\ref{thm:sba}: it adapts per stage to shrink $\Delta$ while regularizing complexity through $c(o)$. The module is thus an efficient, simple-to-reproduce residual matching procedure that improves stability and generalization.

\subsection{Persistent Topology Alignment}
\label{subsec:topo}

\noindent \textbf{Pull-to-Anchor Effect and Persistent Homology.}
Despite DRC, we observe a systematic contraction in the teacher feature space $\phi(\cdot)$, i.e., a \emph{pull-to-anchor} effect. Let $o_i^\star$ denote the retrieved real anchor for a synthetic image $\tilde{x}_i$. A single residual update in Eqn.~\ref{eqn:drc} yields:
\begin{equation}
\tilde{x}_i' \;=\; \alpha\,\tilde{x}_i \;+\; (1-\alpha)\,o_i^\star.
\end{equation}
Writing $y_i=\phi(\tilde{x}_i)$ and $a_i=\phi(o_i^\star)$, a first-order linearization of $\phi$ on the class manifold\footnote{Locally, $\phi(\alpha u+(1-\alpha)v)\approx \alpha\,\phi(u)+(1-\alpha)\,\phi(v)$.} gives the proxy update $y_i' \approx \alpha\,y_i + (1-\alpha)\,a_i$. For any two samples $i,j$,
\begin{equation}
\label{eqn:feature-effect}
\begin{aligned}
\|y_i' - y_j'\|_2
&\approx \big\| \alpha\,(y_i - y_j) + (1-\alpha)\,(a_i - a_j) \big\|_2 \\
&\le \alpha\,\|y_i-y_j\|_2 \;+\; (1-\alpha)\,\|a_i-a_j\|_2,
\end{aligned}
\end{equation}
by the triangle inequality. Because retrieval favors anchors already near the queries in teacher space, the intra-class distance $\|a_i-a_j\|_2$ is typically small. Repeated application of Eqn.~\ref{eqn:feature-effect} therefore contracts intra-class synthetic distances, causing distinct local clusters to merge prematurely.

To characterize this phenomenon, we turn to topological data analysis~\cite{PDA} and persistent homology (PH)~\cite{PH}. Fix a class and form an $\varepsilon$-neighborhood graph over features; as $\varepsilon$ grows, connected components merge and 1D loops emerge. Completing cliques yields the Vietoris-Rips (VR) filtration~\cite{carlsson2009topology}. Let $\mathrm{B}_0(\varepsilon)$ and $\mathrm{B}_1(\varepsilon)$ denote Betti curves for connected components and $1$-cycles, respectively. The pull-to-anchor effect appears as a \emph{left shift} of $\mathrm{B}_0^{\text{syn}}$ (components merge at smaller $\varepsilon$ than in real data) and a \emph{suppression} of $\mathrm{B}_1^{\text{syn}}$ (loops vanish earlier). We thus summarize the discrepancy for class $c$ by
\begin{equation}
\label{eqn:kappa}
\begin{aligned}
\kappa_c
&= \int_{0}^{\varepsilon_{\max}}\!\big(\mathrm{B}_0^{\mathrm{real}}(\varepsilon)
-\mathrm{B}_0^{\mathrm{syn}}(\varepsilon)\big)_{+}\, d\varepsilon \\
&\quad + \gamma \int_{0}^{\varepsilon_{\max}}\!\big(\mathrm{B}_1^{\mathrm{real}}(\varepsilon)
-\mathrm{B}_1^{\mathrm{syn}}(\varepsilon)\big)_{+}\, d\varepsilon,
\end{aligned}
\end{equation}
with $(x)_{+}=\max(x,0)$, weighting $\gamma>0$, and cutoff $\varepsilon_{\max}$. Larger $\kappa_c$ indicates stronger early merging. Since directly optimizing $\kappa_c$ is non-differentiable, we next introduce a smooth surrogate that aligns real and synthetic topological signatures and thus counters the pull-to-anchor bias.

\noindent \textbf{Persistent Topology Alignment (PTA).}
To counter the contraction predicted by Eqn.~\ref{eqn:feature-effect}, we introduce \emph{Persistent Topology Alignment}. For each class $c$, let
$Z_c^{\mathrm{syn}}=\{\phi(\tilde{x}_i)\}$ be synthetic features and
$Z_c^{\mathrm{real}}=\{\phi(x):x\in p_c\}$ be real features drawn from the same per-class pool $p_c$ used by DRC (including current anchors $a_i$). On the union
$Z_c = Z_c^{\mathrm{syn}} \cup Z_c^{\mathrm{real}}$ with Euclidean metric $d(\cdot,\cdot)$, we approximate a VR \cite{DBLP:books/daglib/0025666} filtration via a \emph{class-balanced mutual $k$-NN} graph: subsample $n_c$ real and $n_c$ synthetic points, connect mutual $k$-nearest neighbors, assign each edge weight $w_{uv}=d(u,v)$, and induce a VR-like filtration by thresholding edges with $d(u,v)\le\varepsilon$ for $\varepsilon\in[0,\varepsilon_{\max}]$ while completing cliques to a flag complex.

Running PH on this filtration yields diagrams $\mathcal{D}_c^{(q)}(Z_c)$ for $q\in\{0,1\}$, whose points $(b_j,d_j)$ record birth and death scales of connected components ($q{=}0$) and loops ($q{=}1$). To obtain a stable, differentiable training signal, we map each diagram to a persistence image (PI) on the birth-persistence plane~\cite{DBLP:journals/jmlr/AdamsEKNPSCHMZ17}. With grid centers $\{u_m\}_{m=1}^{M}$, bandwidth $\sigma>0$, and persistence weights $w_q(p)$, the PI for degree $q$ is:
\begingroup
\small
\begin{equation}\label{eqn:pi-def}
I^{(q)}(Z_c)[m]
\!=\!\!\!\!\!
\sum_{(b_j,p_j)\in\mathcal{D}_c^{(q)}(Z_c)}
\!\!\!\!\!\!w_q(p_j)
\exp\!\Big(-\frac{\|\,u_m-(b_j,p_j)\,\|_2^{2}}{2\sigma^2}\Big),
\end{equation}
\endgroup
where $p_j=d_j-b_j$, and both axes are normalized by $\varepsilon_{\max}$. We then align real and synthetic topology by matching PIs:
\begin{equation}
\label{eqn:ltopo}
\begin{aligned}
\mathcal{L}_{\text{topo}}
= \sum_{c}\Big(
&\|I^{(0)}(Z_c^{\mathrm{syn}})-I^{(0)}(Z_c^{\mathrm{real}})\|_2^{2} \\
&\quad + \gamma\,\|I^{(1)}(Z_c^{\mathrm{syn}})-I^{(1)}(Z_c^{\mathrm{real}})\|_2^{2}
\Big),
\end{aligned}
\end{equation}
with loop weight $\gamma>0$. The overall objective is updated as
$\mathcal{L}\leftarrow\mathcal{L}+\lambda_{\text{topo}}\mathcal{L}_{\text{topo}}$ with hyperparameter $\lambda_{\text{topo}}$.

In practice,  we treat $\phi$ as a frozen teacher: its parameters are fixed, but $\phi$ remains in the computation graph so that gradients from $\mathcal{L}_{\text{topo}}$ flow to the synthetic inputs via $\partial\phi/\partial x$. To control overhead, we maintain a feature cache and refresh $\phi(\tilde{x})$ and $\phi(x)$ every $T$ steps (with gradients enabled for the current synthetic batch), while reusing cached features for retrieval and graph construction between refreshes. We use $k\approx10$-$20$, a $32{\times}32$ PI grid, and apply $\mathcal{L}_{\text{topo}}$ at stage ends or every $T$ steps. This preserves $\phi$ as a stable topological reference while providing end-to-end gradients that align birth-persistence statistics of real and synthetic features,  mitigating the pull-to-anchor effect.

\begin{table*}[t]
  \centering
  
  {
  \setlength{\tabcolsep}{10.2pt}
  \renewcommand{\arraystretch}{1.05}

  % helper for cells without a delta（如果别处要用可以保留，当前表里不用也没关系）
  \newcommand{\nodlt}{\multicolumn{1}{c}{\,}}
  % helper: RETA 单元格（仅数值，灰色底）
  \newcommand{\ourscell}[1]{%
    \multicolumn{2}{c}{\cellcolor{gray!15}\Best{#1}}%
  }
  % helper: RETA 单元格（仅数值，灰色底，右侧带竖线）
  \newcommand{\ourscellb}[1]{%
    \multicolumn{2}{c|}{\cellcolor{gray!15}\Best{#1}}%
  }
  % helper: Delta 单元格（蓝字 + 灰底）
  \newcommand{\deltacell}[1]{%
    \multicolumn{2}{c}{\cellcolor{gray!15}\textcolor{blue}{\footnotesize #1}}%
  }
  % helper: Delta 单元格（蓝字 + 灰底，右侧带竖线）
  \newcommand{\deltacellb}[1]{%
    \multicolumn{2}{c|}{\cellcolor{gray!15}\textcolor{blue}{\footnotesize #1}}%
  }
  
  % helper: second-best 单元格（用下划线标记）
  \newcommand{\secondbest}[1]{\underline{#1}}
  %\caption{Test accuracies (\%) on different datasets for baseline methods and RETA with IPC=1, 10 and 50, across ResNet18, ResNet50 and ResNet101. All experiments adopted the settings used in FADRM: 300 epochs for Tiny-ImageNet (IPC=10 and 50), ImageNet-1K and its subsets (ImageNette and ImageWoof); 1000 epochs for CIFAR-100 and Tiny-ImageNet (IPC=1). Since FADRM+ with multiple models can achieve the best performance of FADRM, we only use FADRM+ for comparison in the table. \textbf{-} denotes results not reported in the original paper or not evaluated. All baseline results are taken from FADRM, and our results for RETA are averaged over 5 independent evaluations. The bold font indicates the highest accuracy, the \underline{underlined} scores denote the second-best results, and $\Delta$ reports the improvement of RETA over the second-best method for each setting. IPC: Images Per Class.}
  \caption{Test accuracy (\%) on CIFAR-$100$, Tiny-ImageNet, ImageNette, ImageWoof, and ImageNet-$1$K under $\text{IPC}\in{1,10,50}$ with ResNet-$18/50/101$. Training follows the FADRM protocol: $300$ epochs for Tiny-ImageNet (IPC=$10/50$), ImageNet-$1$K, and its subsets; $1{,}000$ epochs for CIFAR-$100$ and Tiny-ImageNet (IPC=$1$). We report FADRM+ (the strongest FADRM variant); “--” indicates missing or not-evaluated results. Baseline numbers are taken from FADRM; RETA results are averaged over five runs. \textbf{Bold} denotes the best, \underline{underline} the second-best, and $\Delta$ is RETA’s gain over the second-best. IPC = images per class.}

  \label{tab:mainresults}
  \begin{adjustbox}{width=1.0\linewidth} 
  \begin{tabular}{
    c| % backbone
    l % method
    l % venue
    | *{3}{S[table-format=2.1] @{\;} r @{\quad}} % CIFAR-100
      *{3}{S[table-format=2.1] @{\;} r @{\quad}} % Tiny-ImageNet
      *{3}{S[table-format=2.1] @{\;} r @{\quad}} % ImageNette
      *{3}{S[table-format=2.1] @{\;} r @{\quad}} % ImageWoof
      *{3}{S[table-format=2.1] @{\;} r @{\quad}} % ImageNet-1K
  }
    \toprule
      & \multirow{2}{*}{Method} & \multirow{2}{*}{Venue}
      & \multicolumn{6}{c|}{\textbf{CIFAR-100}}
      & \multicolumn{6}{c|}{\textbf{Tiny-ImageNet}}
      & \multicolumn{6}{c|}{\textbf{ImageNette}}
      & \multicolumn{6}{c|}{\textbf{ImageWoof}}
      & \multicolumn{6}{c}{\textbf{ImageNet-1K}} \\
    \cmidrule(lr){4-9} \cmidrule(lr){10-15} \cmidrule(lr){16-21} \cmidrule(lr){22-27} \cmidrule(lr){28-33}
      &
      &
      & \multicolumn{2}{c}{1} & \multicolumn{2}{c}{10} & \multicolumn{2}{c|}{50}
      & \multicolumn{2}{c}{1} & \multicolumn{2}{c}{10} & \multicolumn{2}{c|}{50}
      & \multicolumn{2}{c}{1} & \multicolumn{2}{c}{10} & \multicolumn{2}{c|}{50}
      & \multicolumn{2}{c}{1} & \multicolumn{2}{c}{10} & \multicolumn{2}{c|}{50}
      & \multicolumn{2}{c}{1} & \multicolumn{2}{c}{10} & \multicolumn{2}{c}{50} \\
    \midrule

    % ===== ResNet18 =====
    \multirow{7}{*}{\rotatebox[origin=c]{270}{\textbf{ResNet18}}}%
    & SRe$^2$L \cite{SRe2L} & \textit{NeurIPS'23}
      & \multicolumn{2}{c}{6.6}  & \multicolumn{2}{c}{27.0} & \multicolumn{2}{c|}{50.2}
      & \multicolumn{2}{c}{2.5}  & \multicolumn{2}{c}{16.1} & \multicolumn{2}{c|}{41.1}
      & \multicolumn{2}{c}{19.1} & \multicolumn{2}{c}{29.4} & \multicolumn{2}{c|}{40.9}
      & \multicolumn{2}{c}{--}   & \multicolumn{2}{c}{--}   & \multicolumn{2}{c|}{--}
      & \multicolumn{2}{c}{0.1}  & \multicolumn{2}{c}{21.3} & \multicolumn{2}{c}{46.8} \\
    & RDED \cite{RDED} & \textit{CVPR'24}
      & \multicolumn{2}{c}{17.1} & \multicolumn{2}{c}{56.9} & \multicolumn{2}{c|}{66.8}
      & \multicolumn{2}{c}{11.8} & \multicolumn{2}{c}{41.9} & \multicolumn{2}{c|}{58.2}
      & \multicolumn{2}{c}{35.8} & \multicolumn{2}{c}{61.4} & \multicolumn{2}{c|}{80.4}
      & \multicolumn{2}{c}{20.8} & \multicolumn{2}{c}{38.5} & \multicolumn{2}{c|}{68.5}
      & \multicolumn{2}{c}{6.6}  & \multicolumn{2}{c}{42.0} & \multicolumn{2}{c}{56.5} \\
    & EDC \cite{EDC} & \textit{NeurIPS'24}
      & \multicolumn{2}{c}{39.7} & \multicolumn{2}{c}{63.7} & \multicolumn{2}{c|}{68.6}
      & \multicolumn{2}{c}{39.2} & \multicolumn{2}{c}{51.2} & \multicolumn{2}{c|}{57.2}
      & \multicolumn{2}{c}{--}   & \multicolumn{2}{c}{--}   & \multicolumn{2}{c|}{--}
      & \multicolumn{2}{c}{--}   & \multicolumn{2}{c}{--}   & \multicolumn{2}{c|}{--}
      & \multicolumn{2}{c}{12.8} & \multicolumn{2}{c}{48.6} & \multicolumn{2}{c}{58.0} \\
    & NRR-DD \cite{NRR-DD} & \textit{CVPR'25}
      & \multicolumn{2}{c}{33.3}  & \multicolumn{2}{c}{62.7} & \multicolumn{2}{c|}{67.1}
      & \multicolumn{2}{c}{13.5}  & \multicolumn{2}{c}{45.2} & \multicolumn{2}{c|}{\secondbest{61.2}}
      & \multicolumn{2}{c}{40.1} & \multicolumn{2}{c}{66.2} & \multicolumn{2}{c|}{\secondbest{85.6}}
      & \multicolumn{2}{c}{--}   & \multicolumn{2}{c}{--}   & \multicolumn{2}{c|}{--}
      & \multicolumn{2}{c}{11.6}  & \multicolumn{2}{c}{46.1} & \multicolumn{2}{c}{60.2} \\
    & CaO$_2$ \cite{Cao2} & \textit{ICCV'25}
      & \multicolumn{2}{c}{--} & \multicolumn{2}{c}{--} & \multicolumn{2}{c|}{--}
      & \multicolumn{2}{c}{--} & \multicolumn{2}{c}{--} & \multicolumn{2}{c|}{--}
      & \multicolumn{2}{c}{\secondbest{40.6}}   & \multicolumn{2}{c}{65.0}   & \multicolumn{2}{c|}{84.5}
      & \multicolumn{2}{c}{21.1}   & \multicolumn{2}{c}{45.6}   & \multicolumn{2}{c|}{68.9}
      & \multicolumn{2}{c}{7.1} & \multicolumn{2}{c}{46.1} & \multicolumn{2}{c}{60.0} \\
    & WMDD \cite{WMDD} & \textit{ICCV'25}
      & \multicolumn{2}{c}{--}  & \multicolumn{2}{c}{--} & \multicolumn{2}{c|}{--}
      & \multicolumn{2}{c}{7.6}  & \multicolumn{2}{c}{41.8} & \multicolumn{2}{c|}{59.4}
      & \multicolumn{2}{c}{40.2} & \multicolumn{2}{c}{64.8} & \multicolumn{2}{c|}{83.5}
      & \multicolumn{2}{c}{--}   & \multicolumn{2}{c}{--}   & \multicolumn{2}{c|}{--}
      & \multicolumn{2}{c}{3.2}  & \multicolumn{2}{c}{38.2} & \multicolumn{2}{c}{57.6} \\
    & FADRM+ \cite{FADRM} & \textit{NeurIPS'25}
      & \multicolumn{2}{c}{\secondbest{40.6}} & \multicolumn{2}{c}{\secondbest{67.9}} & \multicolumn{2}{c|}{\secondbest{71.3}}
      & \multicolumn{2}{c}{\secondbest{40.4}} & \multicolumn{2}{c}{\secondbest{52.8}} & \multicolumn{2}{c|}{58.7}
      & \multicolumn{2}{c}{39.2} & \multicolumn{2}{c}{\secondbest{69.0}} & \multicolumn{2}{c|}{84.6}
      & \multicolumn{2}{c}{\secondbest{22.8}} & \multicolumn{2}{c}{\secondbest{57.3}} & \multicolumn{2}{c|}{\secondbest{72.6}}
      & \multicolumn{2}{c}{\secondbest{14.7}} & \multicolumn{2}{c}{\secondbest{50.9}} & \multicolumn{2}{c}{\secondbest{61.2}} \\  
    & \cellcolor{gray!15}\textbf{RETA} & \cellcolor{gray!15}\textit{Ours}
      & \ourscell{42.4} & \ourscell{70.3} & \ourscellb{73.6}
      & \ourscell{43.7} & \ourscell{56.2} & \ourscellb{61.3}
      & \ourscell{42.7} & \ourscell{72.5} & \ourscellb{87.2}
      & \ourscell{24.7} & \ourscell{60.3} & \ourscellb{74.4}
      & \ourscell{16.8} & \ourscell{53.2} & \ourscell{64.3} \\ \cmidrule(lr){2-33}
    & \cellcolor{gray!15}\textcolor{blue}{$\Delta$} & \cellcolor{gray!15}{}
      & \deltacell{$+1.8$}
      & \deltacell{$+2.4$}
      & \deltacellb{$+2.3$}
      & \deltacell{$+3.3$}
      & \deltacell{$+3.4$}
      & \deltacellb{$+0.1$}
      & \deltacell{$+2.1$}
      & \deltacell{$+3.5$}
      & \deltacellb{$+1.6$}
      & \deltacell{$+1.9$}
      & \deltacell{$+3.0$}
      & \deltacellb{$+1.8$}
      & \deltacell{$+2.1$}
      & \deltacell{$+2.3$}
      & \deltacell{$+3.1$} \\
    \midrule

    % ===== ResNet50 =====
    \multirow{7}{*}{\rotatebox[origin=c]{270}{\textbf{ResNet50}}}%
    & SRe$^2$L \cite{SRe2L} & \textit{NeurIPS'23}
      & \multicolumn{2}{c}{--}   & \multicolumn{2}{c}{22.4} & \multicolumn{2}{c|}{52.8}
      & \multicolumn{2}{c}{--}   & \multicolumn{2}{c}{--}   & \multicolumn{2}{c|}{42.2}
      & \multicolumn{2}{c}{--}   & \multicolumn{2}{c}{--}   & \multicolumn{2}{c|}{--}
      & \multicolumn{2}{c}{--}   & \multicolumn{2}{c}{--}   & \multicolumn{2}{c|}{--}
      & \multicolumn{2}{c}{0.3}  & \multicolumn{2}{c}{28.4} & \multicolumn{2}{c}{55.6} \\
    & RDED \cite{RDED} & \textit{CVPR'24}
      & \multicolumn{2}{c}{10.9} & \multicolumn{2}{c}{41.6} & \multicolumn{2}{c|}{64.0}
      & \multicolumn{2}{c}{8.2}  & \multicolumn{2}{c}{38.4} & \multicolumn{2}{c|}{45.6}
      & \multicolumn{2}{c}{27.0} & \multicolumn{2}{c}{55.0} & \multicolumn{2}{c|}{81.8}
      & \multicolumn{2}{c}{17.8} & \multicolumn{2}{c}{35.2} & \multicolumn{2}{c|}{67.0}
      & \multicolumn{2}{c}{8.0}  & \multicolumn{2}{c}{49.7} & \multicolumn{2}{c}{62.0} \\
    & EDC \cite{EDC} & \textit{NeurIPS'24}
      & \multicolumn{2}{c}{36.1} & \multicolumn{2}{c}{62.1} & \multicolumn{2}{c|}{69.4}
      & \multicolumn{2}{c}{35.9} & \multicolumn{2}{c}{50.2} & \multicolumn{2}{c|}{58.8}
      & \multicolumn{2}{c}{--}   & \multicolumn{2}{c}{--}   & \multicolumn{2}{c|}{--}
      & \multicolumn{2}{c}{--}   & \multicolumn{2}{c}{--}   & \multicolumn{2}{c|}{--}
      & \multicolumn{2}{c}{13.3} & \multicolumn{2}{c}{54.1} & \multicolumn{2}{c}{64.3} \\
    & NRR-DD \cite{NRR-DD} & \textit{CVPR'25}
      & \multicolumn{2}{c}{31.8}  & \multicolumn{2}{c}{61.4} & \multicolumn{2}{c|}{66.3}
      & \multicolumn{2}{c}{8.4}  & \multicolumn{2}{c}{43.7} & \multicolumn{2}{c|}{56.9}
      & \multicolumn{2}{c}{31.1} & \multicolumn{2}{c}{62.3} & \multicolumn{2}{c|}{77.9}
      & \multicolumn{2}{c}{--}   & \multicolumn{2}{c}{--}   & \multicolumn{2}{c|}{--}
      & \multicolumn{2}{c}{9.5}  & \multicolumn{2}{c}{50.2} & \multicolumn{2}{c}{62.2} \\
    & CaO$_2$ \cite{Cao2} & \textit{ICCV'25}
      & \multicolumn{2}{c}{--} & \multicolumn{2}{c}{--} & \multicolumn{2}{c|}{--}
      & \multicolumn{2}{c}{--} & \multicolumn{2}{c}{--} & \multicolumn{2}{c|}{--}
      & \multicolumn{2}{c}{\secondbest{33.5}}   & \multicolumn{2}{c}{67.5}   & \multicolumn{2}{c|}{82.7}
      & \multicolumn{2}{c}{\secondbest{20.6}}   & \multicolumn{2}{c}{40.1}   & \multicolumn{2}{c|}{68.2}
      & \multicolumn{2}{c}{7.0} & \multicolumn{2}{c}{53.0} & \multicolumn{2}{c}{65.5} \\
    & WMDD \cite{WMDD} & \textit{ICCV'25}
      & \multicolumn{2}{c}{--}  & \multicolumn{2}{c}{--} & \multicolumn{2}{c|}{--}
      & \multicolumn{2}{c}{5.2}  & \multicolumn{2}{c}{38.1} & \multicolumn{2}{c|}{53.8}
      & \multicolumn{2}{c}{32.7} & \multicolumn{2}{c}{60.3} & \multicolumn{2}{c|}{78.4}
      & \multicolumn{2}{c}{--}   & \multicolumn{2}{c}{--}   & \multicolumn{2}{c|}{--}
      & \multicolumn{2}{c}{--}  & \multicolumn{2}{c}{--} & \multicolumn{2}{c}{61.2} \\
    & FADRM+ \cite{FADRM} & \textit{NeurIPS'25}
      & \multicolumn{2}{c}{\secondbest{37.4}} & \multicolumn{2}{c}{\secondbest{67.4}} & \multicolumn{2}{c|}{\secondbest{72.1}}
      & \multicolumn{2}{c}{\secondbest{39.4}} & \multicolumn{2}{c}{\secondbest{53.7}} & \multicolumn{2}{c|}{\secondbest{60.3}}
      & \multicolumn{2}{c}{31.9} & \multicolumn{2}{c}{\secondbest{68.1}} & \multicolumn{2}{c|}{\secondbest{85.4}}
      & \multicolumn{2}{c}{19.9} & \multicolumn{2}{c}{\secondbest{54.1}} & \multicolumn{2}{c|}{\secondbest{71.7}}
      & \multicolumn{2}{c}{\secondbest{16.2}} & \multicolumn{2}{c}{\secondbest{57.5}} & \multicolumn{2}{c}{\secondbest{66.9}} \\
    & \cellcolor{gray!15}\textbf{RETA} & \cellcolor{gray!15}\textit{Ours}
      & \ourscell{39.9} & \ourscell{69.2} & \ourscellb{74.2}
      & \ourscell{41.5} & \ourscell{55.8} & \ourscellb{62.8}
      & \ourscell{34.2} & \ourscell{70.2} & \ourscellb{86.9}
      & \ourscell{22.4} & \ourscell{56.5} & \ourscellb{73.6}
      & \ourscell{18.7} & \ourscell{59.6} & \ourscell{68.4} \\
    \cmidrule(lr){2-33}
    & \cellcolor{gray!15}\textcolor{blue}{$\Delta$} & \cellcolor{gray!15}{}
      & \deltacell{$+2.5$}
      & \deltacell{$+1.8$}
      & \deltacellb{$+2.1$}
      & \deltacell{$+2.1$}
      & \deltacell{$+2.1$}
      & \deltacellb{$+2.5$}
      & \deltacell{$+0.7$}
      & \deltacell{$+2.1$}
      & \deltacellb{$+1.5$}
      & \deltacell{$+1.8$}
      & \deltacell{$+2.4$}
      & \deltacellb{$+1.9$}
      & \deltacell{$+2.5$}
      & \deltacell{$+2.1$}
      & \deltacell{$+1.5$} \\
    \midrule

    % ===== ResNet101 =====
    \multirow{7}{*}{\rotatebox[origin=c]{270}{\textbf{ResNet101}}}%
    & SRe$^2$L \cite{SRe2L} & \textit{NeurIPS'23}
      & \multicolumn{2}{c}{6.2}  & \multicolumn{2}{c}{30.7} & \multicolumn{2}{c|}{56.9}
      & \multicolumn{2}{c}{1.9}  & \multicolumn{2}{c}{14.6} & \multicolumn{2}{c|}{42.5}
      & \multicolumn{2}{c}{15.8} & \multicolumn{2}{c}{23.4} & \multicolumn{2}{c|}{36.5}
      & \multicolumn{2}{c}{--}   & \multicolumn{2}{c}{--}   & \multicolumn{2}{c|}{--}
      & \multicolumn{2}{c}{0.6}  & \multicolumn{2}{c}{30.9} & \multicolumn{2}{c}{60.8} \\
    & RDED \cite{RDED} & \textit{CVPR'24}
      & \multicolumn{2}{c}{11.2} & \multicolumn{2}{c}{54.1} & \multicolumn{2}{c|}{67.9}
      & \multicolumn{2}{c}{9.6}  & \multicolumn{2}{c}{22.9} & \multicolumn{2}{c|}{41.2}
      & \multicolumn{2}{c}{25.1} & \multicolumn{2}{c}{54.0} & \multicolumn{2}{c|}{75.0}
      & \multicolumn{2}{c}{19.6} & \multicolumn{2}{c}{31.3} & \multicolumn{2}{c|}{59.1}
      & \multicolumn{2}{c}{5.9}  & \multicolumn{2}{c}{48.3} & \multicolumn{2}{c}{61.2} \\
    & EDC \cite{EDC} & \textit{NeurIPS'24}
      & \multicolumn{2}{c}{32.3} & \multicolumn{2}{c}{61.7} & \multicolumn{2}{c|}{68.5}
      & \multicolumn{2}{c}{40.6} & \multicolumn{2}{c}{51.6} & \multicolumn{2}{c|}{58.6}
      & \multicolumn{2}{c}{--}   & \multicolumn{2}{c}{--}   & \multicolumn{2}{c|}{--}
      & \multicolumn{2}{c}{--}   & \multicolumn{2}{c}{--}   & \multicolumn{2}{c|}{--}
      & \multicolumn{2}{c}{12.2} & \multicolumn{2}{c}{51.7} & \multicolumn{2}{c}{64.9} \\
    & NRR-DD \cite{NRR-DD} & \textit{CVPR'25}
      & \multicolumn{2}{c}{32.9}  & \multicolumn{2}{c}{58.3} & \multicolumn{2}{c|}{65.1}
      & \multicolumn{2}{c}{10.1}  & \multicolumn{2}{c}{26.1} & \multicolumn{2}{c|}{46.2}
      & \multicolumn{2}{c}{28.1} & \multicolumn{2}{c}{56.0} & \multicolumn{2}{c|}{78.0}
      & \multicolumn{2}{c}{--}   & \multicolumn{2}{c}{--}   & \multicolumn{2}{c|}{--}
      & \multicolumn{2}{c}{12.2}  & \multicolumn{2}{c}{51.3} & \multicolumn{2}{c}{64.3} \\
    & CaO$_2$ \cite{Cao2} & \textit{ICCV'25}
      & \multicolumn{2}{c}{--} & \multicolumn{2}{c}{--} & \multicolumn{2}{c|}{--}
      & \multicolumn{2}{c}{--} & \multicolumn{2}{c}{--} & \multicolumn{2}{c|}{--}
      & \multicolumn{2}{c}{\secondbest{32.7}}   & \multicolumn{2}{c}{\secondbest{66.3}}   & \multicolumn{2}{c|}{81.7}
      & \multicolumn{2}{c}{21.2}   & \multicolumn{2}{c}{36.5}   & \multicolumn{2}{c|}{63.1}
      & \multicolumn{2}{c}{6.0} & \multicolumn{2}{c}{52.2} & \multicolumn{2}{c}{66.2} \\
    & WMDD \cite{WMDD} & \textit{ICCV'25}
      & \multicolumn{2}{c}{--}  & \multicolumn{2}{c}{--} & \multicolumn{2}{c|}{--}
      & \multicolumn{2}{c}{4.9}  & \multicolumn{2}{c}{37.9} & \multicolumn{2}{c|}{54.8}
      & \multicolumn{2}{c}{31.7} & \multicolumn{2}{c}{60.2} & \multicolumn{2}{c|}{78.9}
      & \multicolumn{2}{c}{--}   & \multicolumn{2}{c}{--}   & \multicolumn{2}{c|}{--}
      & \multicolumn{2}{c}{--}  & \multicolumn{2}{c}{--} & \multicolumn{2}{c}{62.6} \\  
    & FADRM+ \cite{FADRM} & \textit{NeurIPS'25}
      & \multicolumn{2}{c}{\secondbest{40.1}} & \multicolumn{2}{c}{\secondbest{68.9}} & \multicolumn{2}{c|}{\secondbest{72.1}}
      & \multicolumn{2}{c}{\secondbest{41.9}} & \multicolumn{2}{c}{\secondbest{53.6}} & \multicolumn{2}{c|}{\secondbest{60.8}}
      & \multicolumn{2}{c}{29.3} & \multicolumn{2}{c}{63.7} & \multicolumn{2}{c|}{\secondbest{82.3}}
      & \multicolumn{2}{c}{\secondbest{21.8}} & \multicolumn{2}{c}{\secondbest{51.4}} & \multicolumn{2}{c|}{\secondbest{70.6}}
      & \multicolumn{2}{c}{\secondbest{14.1}} & \multicolumn{2}{c}{\secondbest{58.1}} & \multicolumn{2}{c}{\secondbest{67.0}} \\
    & \cellcolor{gray!15}\textbf{RETA} & \cellcolor{gray!15}\textit{Ours}
      & \ourscell{41.8} & \ourscell{70.4} & \ourscellb{73.9}
      & \ourscell{43.1} & \ourscell{55.6} & \ourscellb{63.4}
      & \ourscell{32.9} & \ourscell{66.7} & \ourscellb{84.1}
      & \ourscell{23.5} & \ourscell{54.6} & \ourscellb{72.2}
      & \ourscell{15.7} & \ourscell{59.2} & \ourscell{67.9} \\
    \cmidrule(lr){2-33}
    & \cellcolor{gray!15}\textcolor{blue}{$\Delta$} & \cellcolor{gray!15}{}
      & \deltacell{$+1.7$}
      & \deltacell{$+1.5$}
      & \deltacellb{$+1.8$}
      & \deltacell{$+1.2$}
      & \deltacell{$+2.0$}
      & \deltacellb{$+2.6$}
      & \deltacell{$+0.2$}
      & \deltacell{$+0.4$}
      & \deltacellb{$+1.8$}
      & \deltacell{$+1.7$}
      & \deltacell{$+3.2$}
      & \deltacellb{$+1.6$}
      & \deltacell{$+1.6$}
      & \deltacell{$+1.1$}
      & \deltacell{$+0.9$} \\
    \bottomrule
  \end{tabular}
  \end{adjustbox}
  }

\end{table*}

\section{Experiment}

In this section, we report our empirical evaluation of \textbf{RETA} in the decoupled dataset distillation against strong state-of-the-art (SOTA) decoupled baselines, and further probe (i) cross-architecture generalization from CNNs to ViTs, (ii) robustness to common corruptions, (iii) computational efficiency, and (iv) the contribution of each module via ablations and qualitative visualizations of the distilled images. 

% For more experiments,  please refer to Appendix \textcolor{red}{B} and \textcolor{red}{C}.

\subsection{Experimental Setup}
% In this section, we detail the experimental setup used to evaluate our proposed method. To ensure a fair and direct comparison with prior works, our evaluation protocol, datasets, and post-evaluation settings strictly follow those established in FADRM.

\noindent\textbf{Datasets.} We evaluate on CIFAR-$100$ ($100$ classes, $32{\times}32$) \cite{CIFAR_10_100}, Tiny-ImageNet (200 classes, $64{\times}64$) \cite{Tiny-ImageNet}, and ImageNet-1K (1,000 classes, $224{\times}224$) \cite{ImageNet-1K}, as well as the ImageNet subsets ImageNette and ImageWoof ($10$ classes each, $224{\times}224$). Unless stated otherwise, results are reported under images-per-class (IPC) , with $\text{IPC}\in{1,10,50}$, to assess performance across compression ratios.

% \textbf{Dataset.} 
% We conduct experiments across several datasets with varying resolutions and scales, including CIFAR-100 (100 classes, 32$\times$32 pixels) \cite{CIFAR_10_100}, Tiny-ImageNet (200 classes, 64$\times$64 pixels) \cite{Tiny-ImageNet} and the large-scale ImageNet-1K (1000 classes, 224$\times$224 pixels) \cite{ImageNet-1K}, along with its subsets ImageNette (10 classes, 224$\times$224 pixels) and ImageWoof (10 classes, 224$\times$224 pixels). Our proposed RETA is performed across multiple Images Per Class (IPC) settings, including IPC=1, 10 and 50, to assess performance under different compression ratios.

\noindent\textbf{Baselines.} We compare RETA to SOTA dataset distillation methods: SRe$^{2}$L~\cite{SRe2L}, RDED~\cite{RDED}, EDC~\cite{EDC}, CaO$_2$~\cite{Cao2}, WMDD~\cite{WMDD}, NRR-DD~\cite{NRR-DD}, and FADRM~\cite{FADRM}. Our primary baseline is FADRM, which employs residual matching to mitigate information vanishing; RETA explicitly addresses the structural limitations of its static residual connection. Following FADRM’s protocol, we evaluate ResNet-$18$/$50$/$101$~\cite{ResNet18}, training each model from scratch on synthetic datasets produced by the competing methods.

% \textbf{Architectures}
% For the main performance evaluation, synthetic datasets are used to train models from scratch . Following the protocol of FADRM , we use ResNet-18, ResNet-50 and ResNet-101 as the primary evaluation architectures. 
% Furthermore, to assess the cross-architecture generalization of the distilled data, we also evaluate on a diverse set of architectures, including EfficientNet-B0, MobileNetV2, Swin-Tiny, DenseNet-121, DenseNet169 and DenseNet201.

\noindent\textbf{Implementation Details.} Unless otherwise noted, we adopt the official FADRM settings~\cite{FADRM}: Adam~\cite{Adam} with learning rate $0.25$ and $(\beta_1,\beta_2){=}(0.5,0.9)$, an optimization budget of $B\in{300,2000}$ steps (dataset-dependent), and a merge ratio $\alpha{=}0.5$. RETA adds two scalars: $\lambda$ for the DRC fit–complexity trade-off and $\lambda_{\text{topo}}$ to weight the PTA loss. All distillation runs are conducted on a single NVIDIA RTX $4090$. Following FADRM, we train under four preset hyperparameter configurations and report the best accuracy. 
% More implementation details are provided in Appendix~\textcolor{red}{A}.  

% \textbf{Implementation details.}
% We adopt the official hyperparameter settings reported in FADRM \cite{FADRM}, including an Adam optimizer \cite{Adam} with a learning rate of 0.25 and betas set to (0.5, 0.9), a total optimization budget B of either 2000 or 300 steps (depending on the dataset), and a merge ratio $\alpha=0.5$. We introduce two main hyperparameters: $\lambda$, which controls the trade-off for the feature-complexity score in the Dynamic Retrieval Connection (DRC), and $\lambda_{\text{topo}}$, which weights the Persistent Topology Alignment (PTA) loss. All distillation experiments are conducted on a single NVIDIA RTX 4090 GPU. Following FADRM, we evaluate models trained under the same four hyperparameter configurations and report the highest achieved accuracy. Additional implementation details and hyperparameter settings are provided in Appendix B.

% \begin{figure}[t]
%   \centering
%   \scalebox{0.94}{
%     \begin{tabular}{c}
%       \vspace{5pt}
%       \hspace{-0.45cm}
%       \includegraphics[width=8.5cm]{Figure/cross_architecture.pdf} \\
%       \vspace{-0.5cm}
%     \end{tabular}
%   }
%   \caption{Test accuracies (\%) on ImageNet-1K with IPC=10 for cross-architecture generalization.}
%   \label{tab:cross-architecture}
% \end{figure}

\begin{table}[t]
  \small
  \centering
      \caption{Test accuracies (\%) on ImageNet-1K with IPC=$10$ for cross-architecture generalization. The bold font indicates the best performance achieved on each architecture.}
    \scalebox{0.85}{%
    \begin{tabular}{lcccccc}
      \toprule
      \multicolumn{1}{l|}{Architecture} & \multicolumn{1}{c|}{\#Params} & RDED & EDC & CV-DD & FADRM+ & \cellcolor{gray!15}\textbf{RETA} \\
      \midrule
      \multicolumn{1}{l|}{EfficientNet-B0} & \multicolumn{1}{c|}{39.6M} & 42.8 & 51.1 & 43.2 & 51.9 & \cellcolor{gray!15}\textbf{53.6} \\ 
      \multicolumn{1}{l|}{MobileNetV2} & \multicolumn{1}{c|}{3.4M} & 34.4 & 45.0 & 39.0 & 45.5 & \cellcolor{gray!15}\textbf{47.4} \\
      \multicolumn{1}{l|}{ShuffleNetV2-0.5x} & \multicolumn{1}{c|}{1.4M} & 19.6 & 29.8 & 27.4 & 30.2 & \cellcolor{gray!15}\textbf{33.6} \\
      \multicolumn{1}{l|}{Swin-Tiny} & \multicolumn{1}{c|}{28.0M} & 29.2 & 38.3 & -- & 39.1 & \cellcolor{gray!15}\textbf{41.5} \\
      \multicolumn{1}{l|}{Wide ResNet50-2} & \multicolumn{1}{c|}{68.9M} & 50.0 & -- & 53.9 & 59.1 & \cellcolor{gray!15}\textbf{60.7} \\
      \multicolumn{1}{l|}{DenseNet121} & \multicolumn{1}{c|}{8.0M} & 49.4 & -- & 50.9 & 55.4 & \cellcolor{gray!15}\textbf{56.8} \\
      \multicolumn{1}{l|}{DenseNet169} & \multicolumn{1}{c|}{14.2M} & 50.9 & -- & 53.6 & 58.5 & \cellcolor{gray!15}\textbf{59.4} \\
      \multicolumn{1}{l|}{DenseNet201} & \multicolumn{1}{c|}{20.0M} & 49.0 & -- & 54.8 & 59.7 & \cellcolor{gray!15}\textbf{60.8} \\     
      \bottomrule
    \end{tabular}%
    }

    \label{tab:cross-architecture}
\end{table}

\subsection{Comparison with the State-of-the-art}
\noindent \textbf{Main Results.} Tab.~\ref{tab:mainresults} reports test accuracy on five benchmarks with ResNet-$18$/$50$/$101$ under $\text{IPC}\in{1,10,50}$. \textbf{RETA} delivers the best result in every setting and consistently surpasses the strongest decoupled baseline (FADRM+), as reflected by the $\Delta$ row. Gains are most pronounced on smaller backbones and higher-resolution datasets, for example, with ResNet-$18$ we observe $+3.5$ points on ImageNette (IPC=$1$/$10$), $+3.4$ points on Tiny-ImageNet (IPC=$10$), and $+3.1$ points on ImageNet-1K (IPC=$50$). Improvements remain robust in the data-scarce regime (IPC=$1$) across all backbones, and persist with deeper networks (typically $+0.9$-$3.2$ points with ResNet-$101$). These trends indicate that retrieval and topology alignment yield more informative synthetic sets, translating into consistent accuracy gains from small to large scale.

\noindent \textbf{Cross-Architecture Generalization.} 
Following FADRM~\cite{FADRM}, we train diverse students from scratch on distilled sets for ImageNet-$1$K with \(\text{IPC}{=}10\): lightweight CNNs (MobileNetV$2$, ShuffleNetV$2$-$0.5$x), larger CNNs (EfficientNet-B$0$, Wide-ResNet$50$-$2$, DenseNet$121/169/201$), and a ViT (Swin-Tiny). As summarized in Tab.~\ref{tab:cross-architecture}, \textbf{RETA} attains the best accuracy on \emph{every} architecture, with gains over the strongest baseline (FADRM+) ranging from \(+0.9\) to \(+3.4\) points (largest on ShuffleNetV2). Improvements hold for both convolutional and transformer paradigms (e.g., Swin-Tiny), and persist as capacity scales from mobile to deep/wide models. These trends suggest that RETA’s distilled images encode architecture-agnostic, transferable cues, i.e., DRC reduces bias by adaptively filtering anchors via a fit–complexity score, while PTA preserves multi-scale class geometry, thereby mitigating overfitting to any single inductive bias.

% \subsection{Cross-Architecture Generalization}
% Following the settings of FADRM \cite{FADRM}, we evaluate the cross-architecture generalization of RETA. Specifically, we train a wide range of student models from scratch using synthetic datasets, including lightweight CNNs \cite{ConvNet} (e.g., MobileNetV2 \cite{MobileNetV2}, ShuffleNetV2 \cite{ShuffleNet}), large-scale CNNs (e.g., EfficientNet-B0 \cite{EfficientNetV2}, Wide ResNet50-2 \cite{ResNet18}, DenseNet series \cite{DenseNet}), and a vision transformer Swin-Tiny \cite{SwinTiny}. We evaluate this capability on the ImageNet-1K with IPC=10, and the results are presented in Tab.~\ref{tab:cross-architecture}. The results demonstrate that our method consistently achieves SOTA performance, outperforming all baselines across every evaluated architecture. This generalization across models with differing paradigms and scales indicates that our distilled dataset encodes more fundamental and transferable visual representations. Consequently, it mitigates overfitting to any particular architectural inductive bias.

\begin{figure}[t]
  \centering
  \scalebox{0.95}{
    \begin{tabular}{c}
      % 略微缩窄 & 去掉过大的左移
      \includegraphics[width=0.95\linewidth]{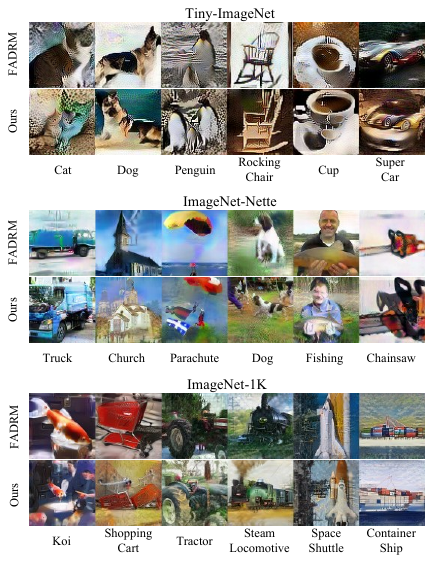} \\
    \end{tabular}
  }
  \caption{Visualization of images distilled by FADRM and RETA. The synthetic images on  Tiny-ImageNet, ImageNette and ImageNet-$1$K are demonstrated here.}
  \label{fig:visualization}
\end{figure}

\noindent \textbf{Visualization of Synthetic Images.}
Fig.~\ref{fig:visualization} contrasts distilled images from FADRM and \textbf{RETA}. Owing to its \emph{static} residual patches, FADRM tends to produce visually tidy but uniform prototypes with reduced intra-class diversity (similar poses/backgrounds and attenuated high-frequency detail). In contrast, \textbf{RETA} yields richer structural variation distinct viewpoints, textures, and layouts without sacrificing class semantics. This diversity is a direct effect of \emph{DRC}, which adaptively selects real patches via a fit–complexity score, preventing anchor overuse and capturing a broader slice of the data manifold. Complementarily, \emph{PTA} regularizes the global feature geometry, discouraging pull-to-anchor collapse and preserving fine details. The resulting images reflect a more faithful, topologically diverse synthesis, consistent with RETA’s stronger generalization observed across architectures and under corruptions.

\subsection{Ablation Study}
\noindent\textbf{Component Combination Evaluation.} We ablate the two modules of \textbf{RETA}: Dynamic Retrieval Connection (DRC) and Persistent Topology Alignment (PTA), under $\text{IPC}{=}10$ with ResNet-$18$~\cite{ResNet18} on CIFAR-$100$~\cite{CIFAR_10_100}, Tiny-ImageNet~\cite{Tiny-ImageNet}, ImageNette~\cite{ImageNet_Subsets}, and ImageNet-$1$K~\cite{ImageNet-1K}, using FADRM as the baseline. As summarized in Tab.~\ref{tab:component_ablation}, adding either DRC or PTA alone yields consistent gains over FADRM across all datasets. The best performance arises when both are enabled, achieving the largest boost on Tiny-ImageNet (e.g., $+3.4$ points). Notably, the combined improvement exceeds the sum of the individual gains, indicating complementarity: \emph{DRC} improves local feature fidelity by adaptively resolving the fit–complexity trade-off via real-patch retrieval, while \emph{PTA} preserves global class geometry by mitigating the pull-to-anchor effect through topological regularization. Together they jointly refine the synthetic set at local and global scales, producing a synergistic advantage unattainable by either component alone.

\begin{figure}[t]
  \centering
  %--- 第一行 ---
  \begin{subfigure}[t]{0.23\textwidth}
    \centering
    \includegraphics[width=\linewidth, height=0.95\linewidth]{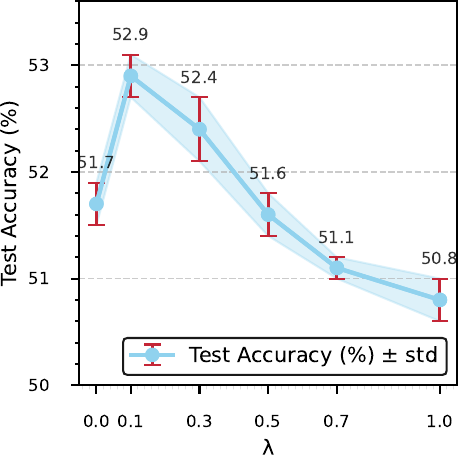}
    \caption{Ablation study for \(\lambda\) in DRC.}
    \label{fig:ablation_lambda}
  \end{subfigure}
  \hfill
  \begin{subfigure}[t]{0.23\textwidth}
    \centering
    \includegraphics[width=\linewidth, height=0.95\linewidth]{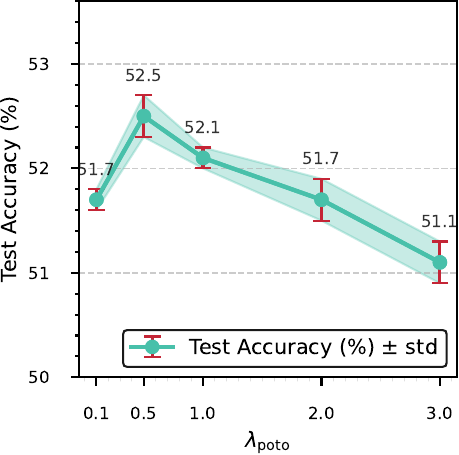}
    \caption{Ablation study for $\lambda_\text{topo}$ in PTA.}
    \label{fig:ablation_lambda_topo}
  \end{subfigure}
  \caption{Ablation studies for hyperparameters on ImageNet-$1$K. Curves show the mean test accuracy, and the error bars denote one standard deviation over five independent runs.}
\end{figure}

\noindent\textbf{Ablation on $\lambda$ in DRC.} We study the trade-off coefficient $\lambda$ in the DRC retrieval score (fit vs. complexity) on ImageNet-$1$K~\cite{ImageNet-1K} with $\text{IPC}{=}10$, sweeping $\lambda\!\in\![0,1]$ while keeping other settings fixed. As shown in Fig.~\ref{fig:ablation_lambda}, accuracy exhibits a clear interior optimum: it rises from $51.7\%$ at $\lambda{=}0$ (no complexity control) to a peak of $52.9\%$ at $\lambda{=}0.1$, then decreases to $50.8\%$ at $\lambda{=}1.0$. Small $\lambda$ fails to penalize ill-conditioned patches, leaving the fit–complexity mismatch unresolved; overly large $\lambda$ over-prioritizes simplicity, discarding informative patches and undermining feature matching. We therefore adopt $\lambda{=}0.1$ as the default setting.

\begin{table}[t]
  \small
  \centering
    \caption{Ablation study on the contributions of different components in our method. The experiments are conducted on CIFAR-$100$, Tiny-ImageNet, ImageNette and ImageNet-$1$K, with IPC=$10$ for all datasets using ResNet-$18$. All results are obtained under optimal hyperparameter settings, and improvements are marked with blue numbers in parentheses.}
  \scalebox{0.8}{
    \begin{tabular}{@{\hskip 0.09in}c@{\hskip 0.09in}c|@{\hskip 0.09in}c|@{\hskip 0.09in}c|@{\hskip 0.09in}c|@{\hskip 0.09in}c}
      \toprule
\textbf{DRC} & \textbf{PTA} & \textbf{CIFAR-100} & \textbf{Tiny-ImageNet} & \textbf{ImageNette}& \textbf{ImageNet-1K} \\
      \midrule
      × & ×           & 67.9 & 52.8 &69.0 & 50.9 \\
      \checkmark & ×  & 69.0{\scriptsize(\dlt{+1.1})} & 54.5{\scriptsize(\dlt{+1.7})} & 70.9{\scriptsize(\dlt{+1.9})} & 51.8{\scriptsize(\dlt{+0.9})} \\
      × & \checkmark  & 68.5{\scriptsize(\dlt{+0.6})} & 53.7{\scriptsize(\dlt{+0.9})} & 69.8{\scriptsize(\dlt{+0.8})}& 51.6{\scriptsize(\dlt{+0.7})} \\
      \cellcolor{gray!10}\checkmark & \cellcolor{gray!10}\checkmark
                                   & \cellcolor{gray!10}{\textbf{70.3}{\scriptsize(\dlt{+2.4})}}
                                   & \cellcolor{gray!10}{\textbf{56.2}{\scriptsize(\dlt{+3.4})}}
                                   & \cellcolor{gray!10}{\textbf{72.5}{\scriptsize(\dlt{+3.5})}} 
                                   & \cellcolor{gray!10}{\textbf{53.2}{\scriptsize(\dlt{+2.3})}} \\
      \bottomrule
    \end{tabular}}

  \label{tab:component_ablation}
\end{table}
\noindent\textbf{Ablation on $\lambda_{\text{topo}}$ in PTA.} We sweep the PTA weight $\lambda_{\text{topo}}\in[0.1,3.0]$ on ImageNet-1K with $\text{IPC}{=}10$, holding other settings fixed. As shown in Fig.~\ref{fig:ablation_lambda_topo}, accuracy exhibits an interior optimum: it increases from $51.7\%$ at $\lambda_{\text{topo}}{=}0.1$ to a peak of $52.5\%$ at $\lambda_{\text{topo}}{=}0.5$, then declines to $51.1\%$ at $\lambda_{\text{topo}}{=}3.0$. With too small a weight, the pull-to-anchor effect is only weakly counteracted and the primary distillation loss dominates. Past the optimum, $\mathcal{L}_{\text{topo}}$ gradients over-regularize geometry and compete with feature-matching objectives, harming fidelity. In our implementation, we therefore adopt $\lambda_{\text{topo}}{=}0.5$ as default.

% \textbf{Ablation Study on hyperparameter $\lambda_\text{topo}$ in PTA.}
% We analyze the impact of the PTA loss weight, $\lambda_\text{topo}$. For this experiment on ImageNet-1K \cite{ImageNet-1K} with IPC=10, we vary $\lambda_\text{topo}$ from 0.1 to 3.0 while keeping other hyperparameters optimal. The results in Figure~\ref{fig:ablation_lambda_topo} show performance rising from 51.7\% at $\lambda_\text{topo}=0.1$ to a peak of 52.5\% at $\lambda_\text{topo}=0.5$, before consistently declining to 51.1\% ($\lambda_\text{topo}=3.0$). Notably, when the weight is minimal ($\lambda_\text{topo}=0.1$), the pull-to-anchor effect we identified is only partially mitigated, as the optimization is still dominated by the primary distillation objectives. The optimal weight ($\lambda_\text{topo}=0.5$) achieves an equilibrium where the topological correction is strong enough to prevent geometric collapse without conflicting with the main loss. Beyond this point ($\lambda_\text{topo} \ge 1.0$), the gradients from $\mathcal{L}_{topo}$ become overly aggressive, competing with the primary objectives and forcing the optimization into a suboptimal state that over-corrects for shape at the expense of core feature fidelity.

\subsection{Computational Efficiency}

\begin{figure}[t]
  \centering
  \scalebox{0.90}{
    \begin{tabular}{c}
      \vspace{5pt}
      \hspace{-0.45cm}
      \includegraphics[width=8.5cm]{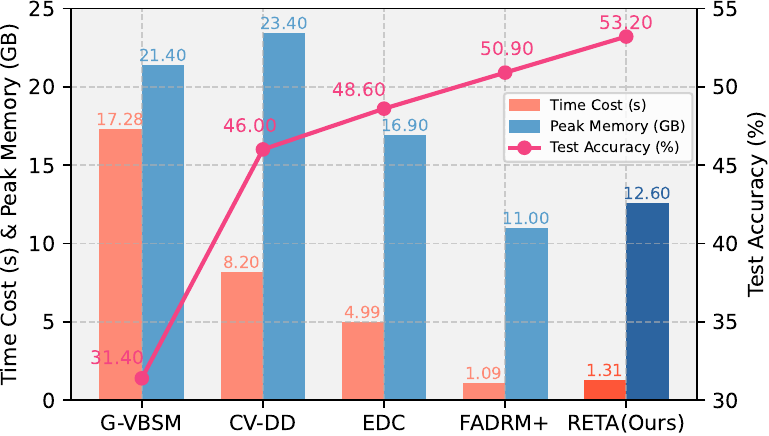} \\
      \vspace{-0.5cm}
    \end{tabular}
  }
  \caption{Computational efficiency comparison between baseline methods and RETA when distilling ImageNet-$1$k. The time cost is measured in seconds, representing the duration required to generate a single image on a single RTX $4090$ GPU.}
  \label{fig:efficiency}
\end{figure}

On ImageNet-$1$K, Fig.~\ref{fig:efficiency} compares wall-clock time and peak GPU memory. \textbf{RETA} remains highly competitive, running substantially lighter than heavier decoupled baselines such as G-VBSM \cite{G-VBSM}, EDC \cite{EDC} (\(4.99\text{s}\), \(17.9,\text{GB}\)) and CV-DD \cite{CV-DD} (\(8.20\text{s}\), \(23.4\text{GB}\)). Relative to the primary baseline FADRM+, RETA introduces only a modest overhead, e.g., \(1.31\text{s}\) vs.\ \(1.09\text{s}\) and \(13.4\text{GB}\) vs.\ \(11.0,\text{GB}\), attributable to the additional computations in DRC and PTA. Given the consistent accuracy gains in Tab.~\ref{tab:mainresults} (e.g., \(+2.3\) points on ImageNet-$1$K at \(\text{IPC}{=}10\)), this constitutes a favorable cost–accuracy trade-off and a practical operating point for large-scale dataset distillation.

% \subsection{Computational Efficiency Comparison.}
% We evaluate the computational efficiency of our method RETA against baseline methods on ImageNet-1K \cite{ImageNet-1K}, with results in Figure~\ref{fig:efficiency}. RETA remains highly competitive, demonstrating significantly lower time cost and peak memory usage compared to most baselines like EDC \cite{EDC} (4.99s, 17.9 GB) and CV-DD \cite{CV-DD}(8.20s, 23.4 GB). When compared to our primary baseline FADRM+ \cite{FADRM} (1.09s, 11.0 GB), our method introduces a modest and expected overhead in both time (1.31s) and memory (13.4 GB). This slight increase is attributable to the necessary computations of our DRC and PTA components, which are required to solve the structural limitations of FADRM. Given the substantial and consistent accuracy gains demonstrated in Tab.~\ref{tab:mainresults} (e.g., +2.3\% on ImageNet-1K  with IPC=10), this modest computational cost represents a highly effective balance for achieving new SOTA performance.

\subsection{Robustness to Corruptions}
We assess robustness under distribution shift by training a ResNet-18 on distilled sets with \(\text{IPC}=1\) and evaluating on ImageNet-Subset-C, averaging over \(15\) corruption types and \(5\) severities. As reported in Tab.~\ref{tab:robustness}, \textbf{RETA} achieves the best accuracy on \emph{all} corrupted subsets (ImageNette-C, ImageWoof-C, ImageFruit-C, ImageYellow-C, ImageMeow-C, ImageSquawk-C), outperforming MTT~\cite{DataMatching_MTT}, IDC~\cite{ParamDistill_IDC}, and the strong decoupled baseline FADRM+~\cite{FADRM}. Relative to FADRM+, the gains are consistent and sizable (mean improvement about \(+3.2\) points), with the largest margins on ImageFruit-C and ImageYellow-C.
We deem that the improvements arise from RETA’s two components acting at complementary scales. \emph{DRC} adaptively filters candidate real patches via a fit–complexity score, avoiding noisy or overly complex injections that static residual links may introduce,  improving generalization under corruption. \emph{PTA} aligns multi-scale class geometry and mitigates the pull-to-anchor effect, yielding less brittle feature structure and stronger invariances to low-level perturbations. Together, these mechanisms produce synthetic images that encode more stable, transferable cues, contributing to  superior performance.

\subsection{Practical Application}
\begin{figure}[t]
  \centering
  \scalebox{0.9}{
    \begin{tabular}{c}
      \includegraphics[width=0.8\linewidth]{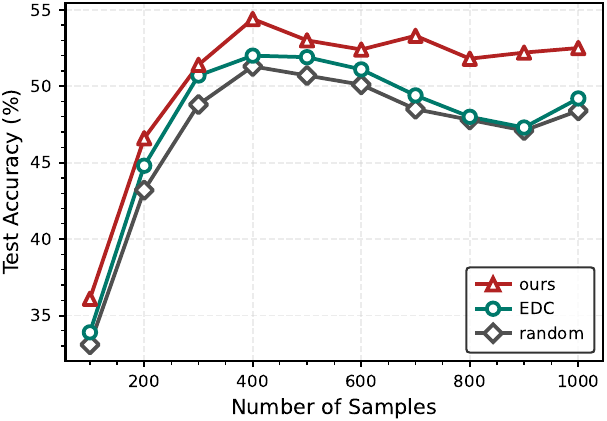} \\
    \end{tabular}
  }
  \caption{Application of continual learning on ImageNet-1k.}
  \label{fig:application}
\end{figure}

To assess the practical utility of our distilled dataset, we evaluate RETA in a challenging class-incremental continual learning (CL) setting~\cite{ContinualLearning_1, ContinualLearning_2, ContinualLearning_3} on ImageNet-$1$K, where a single model sequentially learns disjoint class groups under a fixed replay budget. The distilled images are used as the memory buffer, and we vary the total number of stored samples from $200$ to $1000$.
As shown in Fig.~\ref{fig:application}, RETA consistently achieves higher test accuracy than both the EDC-based buffer and a random real-image buffer across all memory budgets. The margin is already noticeable in the low-budget regime ($200$--$400$ samples), indicating that our distilled images are more sample-efficient replay exemplars. As the buffer grows, the performance of EDC and random selection quickly saturates and even slightly degrades, suggesting redundancy and insufficient structural coverage. In contrast, RETA maintains a stable advantage and continues to improve, demonstrating that the dynamically retrieved and topology-aligned synthetic set captures class structure in a way that better mitigates catastrophic forgetting under realistic memory constraints.

% \subsection{Application}
% To validate the practical utility of our synthetic dataset, we evaluate its effectiveness in a challenging downstream application: class-incremental continual learning (CL) \cite{ContinualLearning_1, ContinualLearning_2, ContinualLearning_3} on ImageNet-1K. This task directly assesses the dataset’s suitability as a memory buffer for mitigating catastrophic forgetting. As shown in Fig.~\ref{fig:application}, our method RETA sustains a consistent and significant accuracy lead over both EDC and FADRM. This superior performance indicates that our method creates a more structurally representative dataset that serves as a more effective buffer against catastrophic forgetting.

\begin{table}[t]
  \centering
  \setlength{\tabcolsep}{5.0pt}
  \renewcommand{\arraystretch}{1.2}
   \caption{Accuracy (\%) on ImageNet-Subset-C with \(\text{IPC}{=}1\). Each score averages over $15$ corruption types and $5$ severities. \textbf{Bold} denotes the best result for each subset.}
  \scalebox{0.870}{
  {\small
  \begin{tabular}{l|cccc}
      \toprule
      \textbf{Datasets} &
      MTT \cite{DataMatching_MTT} &
      IDC \cite{ParamDistill_IDC} &
      FADRM+ \cite{FADRM} &
      \cellcolor{gray!10}\textbf{RETA} \\
      \midrule
      \textbf{ImageNette-C} &
        28.0±1.6 &
        34.5±0.6 &
        37.4±0.4 &
        \cellcolor{gray!10}{\textbf{40.1±0.8}} \\
      \textbf{ImageWoof-C} &
        20.8±1.0 &
        18.7±0.4 &
        19.9±0.7 &
        \cellcolor{gray!10}{\textbf{22.5±0.5}} \\
      \textbf{ImageFruit-C} &
        22.7±1.1 &
        28.5±0.9 &
        30.3±0.8 &
        \cellcolor{gray!10}{\textbf{34.4±0.6}} \\
      \textbf{ImageYellow-C} &
        25.6±1.7 &
        36.8±1.4 &
        37.1±1.3 &
        \cellcolor{gray!10}{\textbf{41.3±1.1}} \\
      \textbf{ImageMeow-C} &
        23.2±1.1 &
        22.2±1.2 &
        25.2±1.1 &
        \cellcolor{gray!10}{\textbf{28.2±0.6}} \\
      \textbf{ImageSquawk-C} &
        25.7±0.8 &
        26.8±0.5 &
        27.3±0.6 &
        \cellcolor{gray!10}{\textbf{29.9±0.5}} \\
      \bottomrule
    \end{tabular}
  }}

  \label{tab:robustness}
\end{table} 

\section{Conclusion}
We studied decoupled dataset distillation with residual matching and showed that \emph{static} anchors induce two coupled failures: a fit–complexity gap and a pull-to-anchor effect,  eroding intra-class diversity and weaken generalization. We introduced \textbf{RETA}, which combines \emph{Dynamic Retrieval Connection} (DRC) to adaptively inject real patches via a fit–complexity score, and \emph{Persistent Topology Alignment} (PTA) to preserve multi-scale class geometry with a persistent-homology loss. Across multiple datasets, RETA delivers consistent SOTA accuracy under standard IPC budgets, transfers robustly across CNN and ViT students.

% \noindent\textbf{Limitations and future work.}
% RETA relies on a frozen teacher, per-class retrieval pools, and several topology/hypergraph choices~\cite{ouyang2024learn} (e.g., $k$-NN construction, PI grid, $\lambda_{\text{topo}}$). The current complexity proxy in DRC is hand-crafted, and PTA adds modest overhead at distillation time. In future work, we plan to develop lighter or streaming topology surrogates for large-scale settings as well as additional modalities and tasks.

\noindent\textbf{Limitations.}
RETA relies on a frozen teacher, per-class retrieval pools, and several topology/hypergraph choices~\cite{ouyang2024learn,DBLP:journals/corr/abs-2508-12379} (e.g., $k$-NN construction, PI grid, $\lambda_{\text{topo}}$). The current complexity proxy in DRC is hand-crafted, and PTA adds modest overhead at distillation time.

\section*{Acknowledgments}
This research was partially supported by the National
Natural Science Foundation of China (NSFC) (62306064) and the Sichuan Science and Technology Program (2023ZYD0165,
2024ZDZX0011 and 2024ZHCG0009).  We appreciate all the authors for their
fruitful discussions. In addition, thanks are extended to
anonymous reviewers for their insightful comments and
suggestions.
{
    \small
    \bibliographystyle{ieeenat_fullname}
    \bibliography{reference}

@inproceedings{DeepLearning_1,
  author    = {Sergey Ioffe and
               Christian Szegedy},
  title     = {Batch Normalization: Accelerating Deep Network Training by Reducing
               Internal Covariate Shift},
  booktitle = {{ICML}},
  year      = {2015}
}

@article{UnifiedTransformerSGGHOI,
  author  = {He, Tao and Gao, Lianli and Song, Jingkuan and Li, Yuan-Fang},
  title   = {Toward a Unified Transformer-Based Framework for Scene Graph Generation and Human-Object Interaction Detection},
  journal = {IEEE Transactions on Image Processing},
  year    = {2023}
}

@inproceedings{He2022OpenVocabSGG,
  author    = {Tao He and Lianli Gao and Jingkuan Song and Yuan{-}Fang Li},
  title     = {Towards Open-Vocabulary Scene Graph Generation with Prompt-Based Finetuning},
  booktitle = {ECCV},
  year      = {2022}
}

@article{naveed2025comprehensive,
  title={A comprehensive overview of large language models},
  author={Naveed, Humza and Khan, Asad Ullah and Qiu, Shi and Saqib, Muhammad and Anwar, Saeed and Usman, Muhammad and Akhtar, Naveed and Barnes, Nick and Mian, Ajmal},
  journal={ACM Transactions on Intelligent Systems and Technology},
  year={2025}
}

@inproceedings{WMDD,
  title={Dataset distillation via the wasserstein metric},
  author={Liu, Haoyang and Li, Yijiang and Xing, Tiancheng and Wang, Peiran and Dalal, Vibhu and Li, Luwei and He, Jingrui and Wang, Haohan},
  booktitle={ICCV},
  year={2025}
}

@article{Survey_1,
  author  = {Shiye Lei and
             Dacheng Tao},
  title   = {A Comprehensive Survey of Dataset Distillation},
  journal = {{IEEE} Trans. Pattern Anal. Mach. Intell.},
  year    = {2024}
}

@inproceedings{DELT,
  author       = {Zhiqiang Shen and
                  Ammar Sherif and
                  Zeyuan Yin and
                  Shitong Shao},
  title        = {{DELT:} {A} Simple Diversity-driven EarlyLate Training for Dataset
                  Distillation},
  booktitle    = {{CVPR}},
  year         = {2025}
}

@inproceedings{NRR-DD,
  author       = {Minh{-}Tuan Tran and
                  Trung Le and
                  Xuan{-}May Le and
                  Thanh{-}Toan Do and
                  Dinh Q. Phung},
  title        = {Enhancing Dataset Distillation via Non-Critical Region Refinement},
  booktitle    = {{CVPR}},
  year         = {2025}
}

@inproceedings{Survey_2,
  author    = {Jiahui Geng and
               Zongxiong Chen and
               Yuandou Wang and
               Herbert Woisetschlaeger and
               Sonja Schimmler and
               Ruben Mayer and
               Zhiming Zhao and
               Chunming Rong},
  title     = {A Survey on Dataset Distillation: Approaches, Applications and Future
               Directions},
  booktitle = {{IJCAI}},
  year      = {2023}
}

@article{Survey_3,
  author  = {Ruonan Yu and
             Songhua Liu and
             Xinchao Wang},
  title   = {Dataset Distillation: {A} Comprehensive Review},
  journal = {{IEEE} Trans. Pattern Anal. Mach. Intell.},
  year    = {2024}
}

@article{DatasetDistillation,
  author  = {Tongzhou Wang and
             Jun{-}Yan Zhu and
             Antonio Torralba and
             Alexei A. Efros},
  title   = {Dataset Distillation},
  journal = {CoRR},
  year    = {2018}
}

@inproceedings{DataMatching_GM,
  author    = {Bo Zhao and
               Konda Reddy Mopuri and
               Hakan Bilen},
  title     = {Dataset Condensation with Gradient Matching},
  booktitle = {{ICLR}},
  year      = {2021}
}

@article{carlsson2009topology,
  title={Topology and data},
  author={Carlsson, Gunnar},
  journal={Bulletin of the American Mathematical Society},
  year={2009}
}

@inproceedings{SRe2L,
  author       = {Zeyuan Yin and
                  Eric P. Xing and
                  Zhiqiang Shen},
  title        = {Squeeze, Recover and Relabel: Dataset Condensation at ImageNet Scale
                  From {A} New Perspective},
  booktitle    = {NeurIPS},
  year         = {2023}
}

@inproceedings{G-VBSM,
  author       = {Shitong Shao and
                  Zeyuan Yin and
                  Muxin Zhou and
                  Xindong Zhang and
                  Zhiqiang Shen},
  title        = {Generalized Large-Scale Data Condensation via Various Backbone and
                  Statistical Matching},
  booktitle    = {{CVPR}},
  year         = {2024}
}

@inproceedings{EDC,
  author       = {Shitong Shao and
                  Zikai Zhou and
                  Huanran Chen and
                  Zhiqiang Shen},
  title        = {Elucidating the Design Space of Dataset Condensation},
  booktitle    = {NeurIPS},
  year         = {2024}
}

@inproceedings{RDED,
  author       = {Peng Sun and
                  Bei Shi and
                  Daiwei Yu and
                  Tao Lin},
  title        = {On the Diversity and Realism of Distilled Dataset: An Efficient Dataset
                  Distillation Paradigm},
  booktitle    = {{CVPR}},
  year         = {2024}
}

@article{DBLP:journals/jmlr/AdamsEKNPSCHMZ17,
  author       = {Henry Adams and
                  Tegan Emerson and
                  Michael Kirby and
                  Rachel Neville and
                  Chris Peterson and
                  Patrick D. Shipman and
                  Sofya Chepushtanova and
                  Eric M. Hanson and
                  Francis C. Motta and
                  Lori Ziegelmeier},
  title        = {Persistence Images: {A} Stable Vector Representation of Persistent
                  Homology},
  journal      = {J. Mach. Learn. Res.},
  year         = {2017}
}

@inproceedings{DataMatching_DSA,
  author    = {Bo Zhao and
               Hakan Bilen},
  title     = {Dataset Condensation with Differentiable Siamese Augmentation},
  booktitle = {{ICML}},
  year      = {2021}
}

@inproceedings{DataMatching_DM,
  author    = {Bo Zhao and
               Hakan Bilen},
  title     = {Dataset Condensation with Distribution Matching},
  booktitle = {{WACV}},
  year      = {2023}
}

@inproceedings{DataMatching_NCFM,
  author    = {Shaobo Wang and
               Yicun Yang and
               Zhiyuan Liu and
               Chenghao Sun and
               Xuming Hu and
               Conghui He and
               Linfeng Zhang},
  title     = {Dataset Distillation with Neural Characteristic Function: {A} Minmax
               Perspective},
  booktitle = {{CVPR}},
  year      = {2025}
}

@inproceedings{DataMatching_MTT,
  author    = {George Cazenavette and
               Tongzhou Wang and
               Antonio Torralba and
               Alexei A. Efros and
               Jun{-}Yan Zhu},
  title     = {Dataset Distillation by Matching Training Trajectories},
  booktitle = {{CVPR}},
  year      = {2022}
}

@inproceedings{DataMatching_FTD,
  author    = {Jiawei Du and
               Yidi Jiang and
               Vincent Y. F. Tan and
               Joey Tianyi Zhou and
               Haizhou Li},
  title     = {Minimizing the Accumulated Trajectory Error to Improve Dataset Distillation},
  booktitle = {{CVPR}},
  year      = {2023}
}

@inproceedings{DataMatching_DATM,
  author    = {Ziyao Guo and
               Kai Wang and
               George Cazenavette and
               Hui Li and
               Kaipeng Zhang and
               Yang You},
  title     = {Towards Lossless Dataset Distillation via Difficulty-Aligned Trajectory
               Matching},
  booktitle = {{ICLR}},
  year      = {2024}
}

@inproceedings{ContinualLearning_1,
  author    = {Wojciech Masarczyk and
               Ivona Tautkute},
  title     = {Reducing catastrophic forgetting with learning on synthetic data},
  booktitle = {{CVPR} Workshops},
  year      = {2020}
}

@inproceedings{ContinualLearning_2,
  author    = {Enneng Yang and
               Li Shen and
               Zhenyi Wang and
               Tongliang Liu and
               Guibing Guo},
  title     = {An Efficient Dataset Condensation Plugin and Its Application to Continual
               Learning},
  booktitle = {NeurIPS},
  year      = {2023}
}

@article{ContinualLearning_3,
  author  = {Da Yu and
             Mingyi Zhang and
             Mantian Li and
             Fusheng Zha and
             Junge Zhang and
             Lining Sun and
             Kaiqi Huang},
  title   = {Contrastive Correlation Preserving Replay for Online Continual Learning},
  journal = {{IEEE} Trans. Circuits Syst. Video Technol.},
  year    = {2024}
}

@inproceedings{ParamDistill_IDC,
  author    = {Jang{-}Hyun Kim and
               Jinuk Kim and
               Seong Joon Oh and
               Sangdoo Yun and
               Hwanjun Song and
               Joonhyun Jeong and
               Jung{-}Woo Ha and
               Hyun Oh Song},
  title     = {Dataset Condensation via Efficient Synthetic-Data Parameterization},
  booktitle = {{ICML}},
  year      = {2022}
}

@inproceedings{ImageNet_Subsets,
  author    = {George Cazenavette and
               Tongzhou Wang and
               Antonio Torralba and
               Alexei A. Efros and
               Jun{-}Yan Zhu},
  title     = {Generalizing Dataset Distillation via Deep Generative Prior},
  booktitle = {{CVPR}},
  year      = {2023}
}

@article{CIFAR_10_100,
  title   = {Learning multiple layers of features from tiny images},
  author  = {Krizhevsky, Alex and Hinton, Geoffrey and others},
  year    = {2009},
  journal = {Technical report Citeseer}
}

@article{Cao2,
  author       = {Haoxuan Wang and
                  Zhenghao Zhao and
                  Junyi Wu and
                  Yuzhang Shang and
                  Gaowen Liu and
                  Yan Yan},
  title        = {CaO\({}_{\mbox{2}}\): Rectifying Inconsistencies in Diffusion-Based
                  Dataset Distillation},
  journal      = {CoRR},
  volume       = {abs/2506.22637},
  year         = {2025}
}

@inproceedings{ResNet18,
  author       = {Kaiming He and
                  Xiangyu Zhang and
                  Shaoqing Ren and
                  Jian Sun},
  title        = {Deep Residual Learning for Image Recognition},
  booktitle    = {{CVPR}},
  year         = {2016}
}

@inproceedings{Adam,
  author       = {Diederik P. Kingma and
                  Jimmy Ba},
  title        = {Adam: {A} Method for Stochastic Optimization},
  booktitle    = {{ICLR}},
  year         = {2015}
}

@inproceedings{MobileNetV2,
  author       = {Mark Sandler and
                  Andrew G. Howard and
                  Menglong Zhu and
                  Andrey Zhmoginov and
                  Liang{-}Chieh Chen},
  title        = {MobileNetV2: Inverted Residuals and Linear Bottlenecks},
  booktitle    = {{CVPR}},
  year         = {2018}
}

@article{CV-DD,
  author       = {Jiacheng Cui and
                  Zhaoyi Li and
                  Xiaochen Ma and
                  Xinyue Bi and
                  Yaxin Luo and
                  Zhiqiang Shen},
  title        = {Dataset Distillation via Committee Voting},
  journal      = {CoRR},
  volume       = {abs/2501.07575},
  year         = {2025}
}

@article{FADRM,
  author       = {Jiacheng Cui and
                  Xinyue Bi and
                  Yaxin Luo and
                  Xiaohan Zhao and
                  Jiacheng Liu and
                  Zhiqiang Shen},
  title        = {{FADRM:} Fast and Accurate Data Residual Matching for Dataset Distillation},
  journal      = {CoRR},
  volume       = {abs/2506.24125},
  year         = {2025}
}

@article{PDA,
  author       = {Fr{\'{e}}d{\'{e}}ric Chazal and
                  Bertrand Michel},
  title        = {An Introduction to Topological Data Analysis: Fundamental and Practical
                  Aspects for Data Scientists},
  journal      = {Frontiers Artif. Intell.},
  year         = {2021}
}

@article{PH,
  author       = {Chi Seng Pun and
                  Si Xian Lee and
                  Kelin Xia},
  title        = {Persistent-homology-based machine learning: a survey and a comparative
                  study},
  journal      = {Artif. Intell. Rev.},
  year         = {2022}
}

@article{Tiny-ImageNet,
  title={Tiny imagenet visual recognition challenge},
  author={Ya Le and Xuan Yang},
  journal={Technical Report},
  year={2015}
}

@article{ImageNet-1K,
  author       = {Olga Russakovsky and
                  Jia Deng and
                  Hao Su and
                  Jonathan Krause and
                  Sanjeev Satheesh and
                  Sean Ma and
                  Zhiheng Huang and
                  Andrej Karpathy and
                  Aditya Khosla and
                  Michael S. Bernstein and
                  Alexander C. Berg and
                  Li Fei{-}Fei},
  title        = {ImageNet Large Scale Visual Recognition Challenge},
  journal      = {Int. J. Comput. Vis.},
  year         = {2015}
}

@inproceedings{DBLP:conf/icml/YangZDR24,
  author       = {William Yang and
                  Ye Zhu and
                  Zhiwei Deng and
                  Olga Russakovsky},
  title        = {What is Dataset Distillation Learning?},
  booktitle    = {{ICML}},
  year         = {2024}
}

@book{DBLP:books/daglib/0025666,
  author       = {Herbert Edelsbrunner and
                  John Harer},
  title        = {Computational Topology - an Introduction},
  publisher    = {American Mathematical Society},
  year         = {2010}
}

@inproceedings{DBLP:conf/mm/LiZ0XLQ24,
  author       = {Muquan Li and
                  Dongyang Zhang and
                  Tao He and
                  Xiurui Xie and
                  Yuan{-}Fang Li and
                  Ke Qin},
  title        = {Towards Effective Data-Free Knowledge Distillation via Diverse Diffusion
                  Augmentation},
  booktitle    = {{ACM MM}},
  year         = {2024}
}

@inproceedings{DBLP:conf/aaai/LiZDXQ25,
  author       = {Muquan Li and
                  Dongyang Zhang and
                  Qiang Dong and
                  Xiurui Xie and
                  Ke Qin},
  title        = {Adaptive Dataset Quantization},
  booktitle    = {AAAI},
  year         = {2025}
}

@inproceedings{libeyond,
  title={Beyond Random: Automatic Inner-loop Optimization in Dataset Distillation},
  author={Li, Muquan and Gou, Hang and Zhang, Dongyang and Liang, Shuang and Xie, Xiurui and Ouyang, Deqiang and Qin, Ke},
  booktitle={NeurIPS},
  year = {2025}
}

@article{ouyang2024learn,
  title={Learn from global correlations: Enhancing evolutionary algorithm via spectral gnn},
  author={Ouyang, Kaichen and Ke, Zong and Fu, Shengwei and Liu, Lingjie and Zhao, Puning and Hu, Dayu},
  journal={arXiv preprint arXiv:2412.17629},
  year={2024}
}

@article{ke2025early,
  title={Early warning of cryptocurrency reversal risks via multi-source data},
  author={Ke, Zong and Cao, Yuqing and Chen, Zhenrui and Yin, Yuchen and He, Shouchao and Cheng, Yu},
  journal={Finance Research Letters},
  year={2025},
}

@article{qiu2025convex,
  title={Convex optimization of Markov decision processes based on Z transform: A theoretical framework for two-space decomposition and linear programming reconstruction},
  author={Qiu, Shiqing and Wang, Haoyu and Zhang, Yuxin and Ke, Zong and Li, Zichao},
  journal={Mathematics},
  year={2025},
}

@article{DBLP:journals/corr/abs-2508-12379,
  author       = {Rongzheng Wang and
                  Qizhi Chen and
                  Yihong Huang and
                  Yizhuo Ma and
                  Muquan Li and
                  Jiakai Li and
                  Ke Qin and
                  Guangchun Luo and
                  Shuang Liang},
  title        = {GraphCogent: Overcoming LLMs' Working Memory Constraints via
                  Multi-Agent Collaboration in Complex Graph Understanding},
  journal      = {CoRR},
  volume       = {abs/2508.12379},
  year         = {2025}
}

@article{wang2026rethinking,
  title={Rethinking LLM-Driven Heuristic Design: Generating Efficient and Specialized Solvers via Dynamics-Aware Optimization},
  author={Wang, Rongzheng and Huang, Yihong and Li, Muquan and Li, Jiakai and Liang, Di and Simons, Bob and Ke, Pei and Liang, Shuang and Qin, Ke},
  journal={arXiv preprint arXiv:2601.20868},
  year={2026}
}

@article{tian2026stpe,
  title={STPE-Map: Multimodal Alignment and Spatio-Temporal Priors for Online HD Map Construction},
  author={Tian, Keke and Li, Muquan and Zhang, Jing and Qin, Ke},
  journal={Knowledge-Based Systems},
  year={2026}
}

@article{li2026efficient,
  title={Efficient Industrial Dataset Distillation With Textual Trajectory Matching},
  author={Li, Muquan and Dong, Qian and Zhang, Dongyang and Qin, Ke and Luo, Guangchun},
  journal={IEEE Transactions on Industrial Informatics},
  year={2026},
  publisher={IEEE}
}

@article{ma2026efficient,
  title={Efficient Dataset Distillation via Generative Pruning},
  author={Ma, Yingyi and Li, Muquan and Duan, Guiduo and Qin, Ke and Liang, Shuang and Zhang, Dongyang},
  journal={IEEE Transactions on Big Data},
  year={2026},
  publisher={IEEE}
}

@article{he2026lifelong,
  title={Lifelong Scene Graph Generation},
  author={He, Tao and Hu, Xin and Wu, Tongtong and Zhang, Dongyang and Li, Ming and Li, Yuan-Fang and Yu, Fei Richard},
  journal={Pattern Recognition},
  pages={113132},
  year={2026}
}

@inproceedings{hu2025spade,
  title={SPADE: Spatial-Aware Denoising Network for Open-vocabulary Panoptic Scene Graph Generation with Long-and Local-range Context Reasoning},
  author={Hu, Xin and Qin, Ke and Duan, Guiduo and Li, Ming and Li, Yuan-Fang and He, Tao},
  booktitle={ICCV},
  year={2025}
}

@ARTICLE{erhc,
  title={Expanding and refining hybrid compressors for efficient object re-identification},
  author={Xie, Yi and Wu, Hanxiao and Zhu, Jianqing and Zeng, Huanqiang and Zhang, Jing},
  journal={IEEE Transactions on Image Processing},
  volume={33},
  pages={3793--3808},
  year={2024}}

@InProceedings{Xie_2024_CVPR,
    author    = {Xie, Yi and Lin, Yihong and Cai, Wenjie and Xu, Xuemiao and Zhang, Huaidong and Du, Yong and He, Shengfeng},
    title     = {D3still: Decoupled Differential Distillation for Asymmetric Image Retrieval},
    booktitle = {CVPR},
    year      = {2024}
}

@inproceedings{cdd,
  title={Towards a Smaller Student: Capacity Dynamic Distillation for Efficient Image Retrieval},
  author={Xie, Yi and Zhang, Huaidong and Xu, Xuemiao and Zhu, Jianqing and He, Shengfeng},
  booktitle={CVPR},
  year={2023}
}

@article{xie2024distillation,
  title={Distillation embedded absorbable pruning for fast object re-identification},
  author={Xie, Yi and Wu, Hanxiao and Zhu, Jianqing and Zeng, Huanqiang},
  journal={Pattern Recognition},
  volume={152},
  pages={110437},
  year={2024}
}

@article{xie2024pairwise,
  title={Pairwise difference relational distillation for object re-identification},
  author={Xie, Yi and Wu, Hanxiao and Lin, Yihong and Zhu, Jianqing and Zeng, Huanqiang},
  journal={Pattern Recognition},
  volume={152},
  pages={110455},
  year={2024}
}

@article{he2026monotonic,
  title={Monotonic Rank Knowledge Distillation via Kendall Correlation},
  author={He, Xuewan and Wang, Jielei and Su, Yuchen and Liu, Dongnan and Zhao, Junbo and Lu, Guoming},
  journal={IEEE Transactions on Circuits and Systems for Video Technology},
  year={2026}
}

@article{he2025prism,
  title={PRISM: Precision-Recall Informed Data-Free Knowledge Distillation via Generative Diffusion},
  author={He, Xuewan and Wang, Jielei and Cheng, Zihan and Su, Yuchen and Huang, Shiyue and Lu, Guoming},
  journal={arXiv preprint arXiv:2509.16897},
  year={2025}
}
}

% WARNING: do not forget to delete the supplementary pages from your submission 
 % \input{sec/X_suppl}

\end{document}